\documentclass{article}
\usepackage[letterpaper, left=1cm, right=1.25cm, top=1cm, bottom=1.25cm]{geometry}
\usepackage{hyperref}
\usepackage{url}
\usepackage[dvips]{graphicx,color}
\usepackage{paralist}
\usepackage{amsmath}
\usepackage{subfigure}
\usepackage{fancyhdr}
\title{Approximate Lesion Localization in Dermoscopy Images}
\author{%
        M. Emre Celebi\\
        Dept. of Computer Science\\Louisiana State Univ., Shreveport, LA, USA\\
        \href{mailto:ecelebi@lsus.edu}{ecelebi@lsus.edu}
\and
        Hitoshi Iyatomi\\
        Dept. of Electrical Informatics\\Hosei Univ., Tokyo, Japan\\
        \href{mailto:iyatomi@hosei.ac.jp}{iyatomi@hosei.ac.jp}
\and
        Gerald Schaefer\\
        School of Engineering and Applied Science\\Aston Univ., Birmingham, UK\\
        \href{mailto:g.schaefer@aston.ac.uk}{g.schaefer@aston.ac.uk}
\and
        William V. Stoecker\\
        Stoecker \& Associates, Rolla, MO, USA\\
        \href{mailto:wvs@mst.edu}{wvs@mst.edu}
       }
\DeclareMathOperator{\I}{I}
\DeclareMathOperator{\U}{U}

\begin{document}


\maketitle
\begin{abstract}

\textbf{Background:} Dermoscopy is one of the major imaging modalities used in the diagnosis of melanoma and other pigmented skin lesions. Due to the difficulty and subjectivity of human interpretation, automated analysis of dermoscopy images has become an important research area. Border detection is often the first step in this analysis. \textbf{Methods:} In this article, we present an approximate lesion localization method that serves as a preprocessing step for detecting borders in dermoscopy images. In this method, first the black frame around the image is removed using an iterative algorithm. The approximate location of the lesion is then determined using an ensemble of thresholding algorithms. \textbf{Results:} The method is tested on a set of 428 dermoscopy images. The localization error is quantified by a metric that uses dermatologist determined borders as the ground truth. \textbf{Conclusion:} The results demonstrate that the method presented here achieves both fast and accurate localization of lesions in dermoscopy images.

\end{abstract}

\section{Introduction}

Malignant melanoma, the most deadly form of skin cancer, is one of the most rapidly increasing cancers in the world, with an estimated incidence of 62,480 and an estimated total of 8,420 deaths in the United States in 2008 alone \cite{Jemal07}. Early diagnosis is particularly important since melanoma can be cured with a simple excision if detected early.
\par
Dermoscopy, also known as epiluminescence microscopy, has become one of the most important tools in the diagnosis of melanoma and other pigmented skin lesions. This non-invasive skin imaging technique involves optical magnification, which makes subsurface structures more easily visible when compared to conventional clinical images \cite{Argenziano02}. This in turn reduces screening errors and provides greater differentiation between difficult lesions such as pigmented Spitz nevi and small, clinically equivocal lesions \cite{Steiner93}. However, it has also been demonstrated that dermoscopy may actually lower the diagnostic accuracy in the hands of inexperienced dermatologists \cite{Binder95}. Therefore, in order to minimize the diagnostic errors that result from the difficulty and subjectivity of visual interpretation, the development of computerized image analysis techniques is of paramount importance \cite{Fleming98}.
\par
Automated border detection is often the first step in the automated or semi-automated analysis of dermoscopy images \cite{Schmid99,Erkol05,Iyatomi06,Celebi07,Celebi08a,Zhou09,Celebi09}. It is crucial for the image analysis for two main reasons. First, the border structure provides important information for accurate diagnosis, as many clinical features, such as asymmetry, border irregularity, and abrupt border cutoff, are calculated directly from the border. Second, the extraction of other important clinical features such as atypical pigment networks, globules, and blue-white areas, critically depends on the accuracy of border detection. Automated border detection is a challenging task due to several reasons:
\begin{inparaenum}[(i)]
\item low contrast between the lesion and the surrounding skin,
\item irregular and fuzzy lesion borders,
\item artifacts such as black frames, skin lines, hairs, and air bubbles,
\item variegated coloring inside the lesion.
\end{inparaenum}
\par
A number of methods have been developed for preprocessing dermoscopy images. Most of these focused on the removal of artifacts such as hairs and bubbles. Of the studies dealing with hair removal, Lee \emph{et al.} \cite{Lee97} and Schmid \cite{Schmid99} approached the problem using mathematical morphology. Fleming \emph{et al.} \cite{Fleming98} applied curvilinear structure detection with various constraints followed by gap filling. More recently, Zhou \emph{et al.} \cite{Zhou08} improved Fleming \emph{et al.}'s approach using feature guided examplar-based inpainting. A method for bubble removal was introduced in \cite{Fleming98}, where the authors utilized a morphological top-hat operator followed by a radial search procedure.
\par
In this article, we present a method for approximate lesion localization in dermoscopy images. First, the black frame around the image is removed using an iterative algorithm. Then, the approximate location of the lesion is determined using an ensemble of thresholding algorithms.

\section{Materials and Methods}
\subsection{Black Frame Removal}
\label{frame_removal}
Dermoscopy images often contain black frames that are introduced during the digitization process. These need to be removed because they
might interfere with the subsequent lesion localization procedure. In order to determine the darkness of a pixel with (R, G, B) coordinates, the lightness component of the HSL color space \cite{Levkowitz93} is utilized:
\begin{equation}
L = \frac{ \max(R,G,B) + \min(R,G,B) }{2}
\end{equation}

\begin{figure}[!ht]
\centering
 \subfigure[Original image]{\label{black_frame_a}\includegraphics[width=0.4\columnwidth]{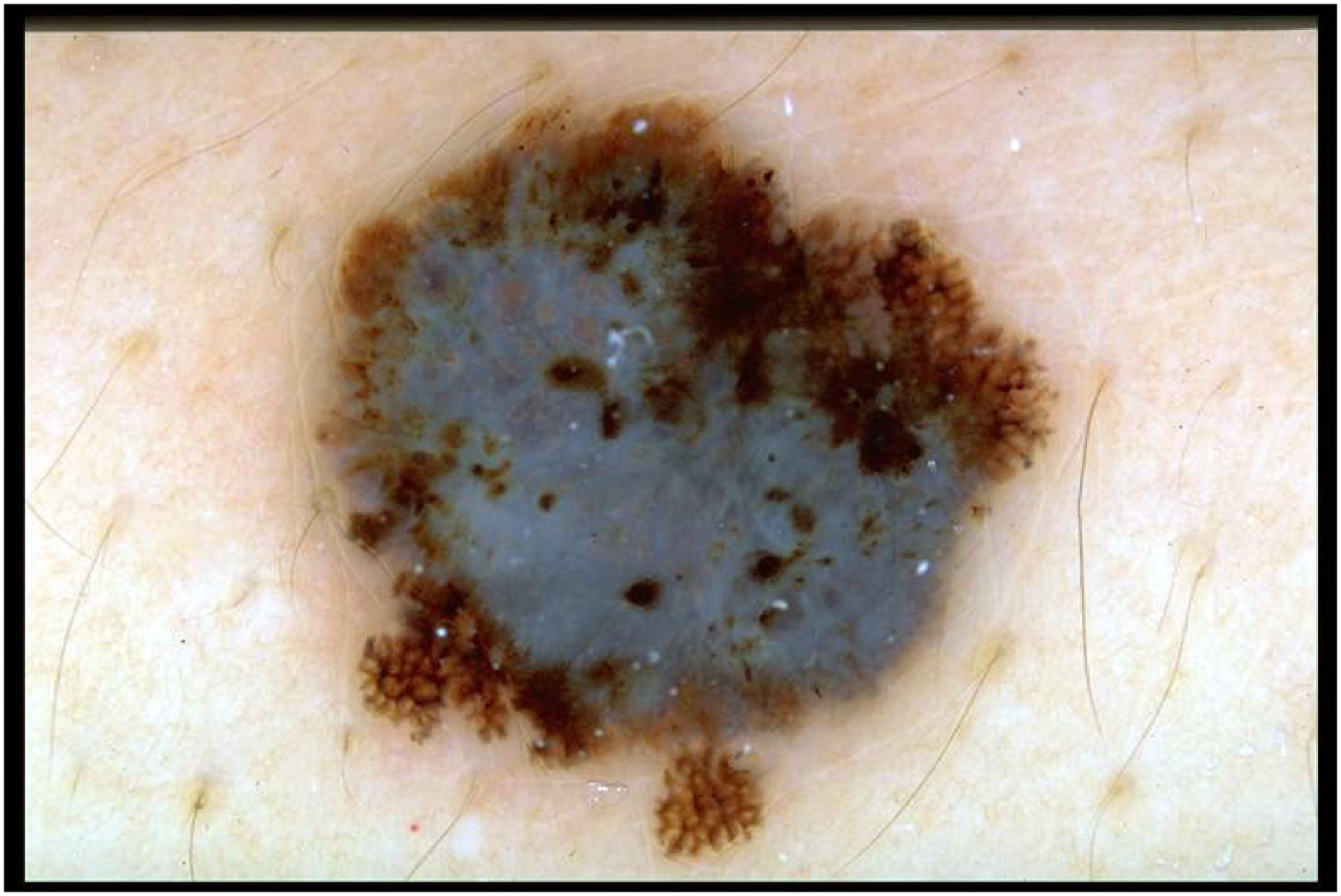}}
 \hspace{.2in}
 \subfigure[After frame removal]{\label{black_frame_b}\includegraphics[width=0.4\columnwidth]{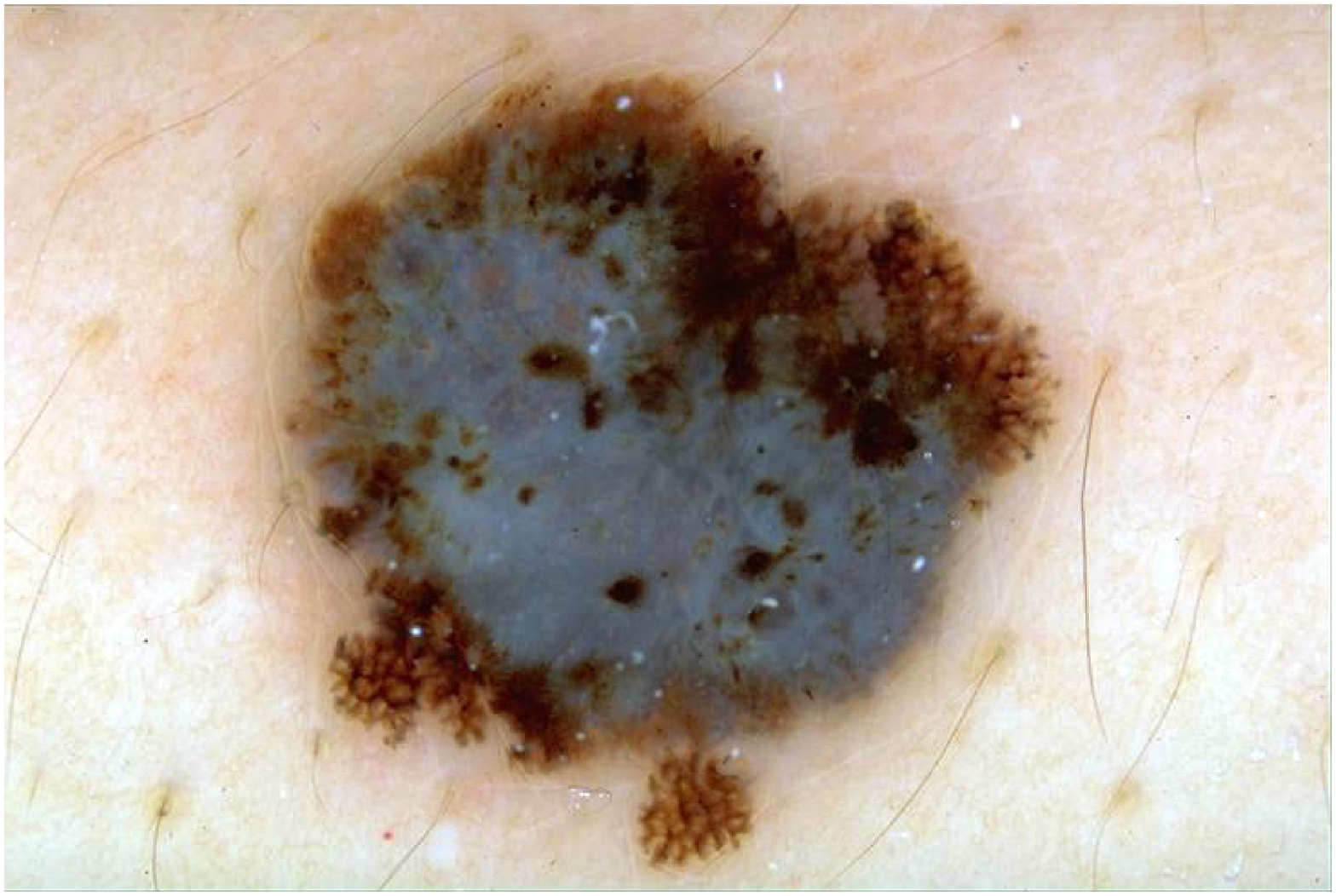}}
 \caption{Black frame removal}
 \label{black_frame}
\end{figure}

In particular, a pixel is considered to be black if its lightness value is less than 20. Using this criterion, the image is scanned row-by-row starting from the top. A particular row is labeled as part of the black frame if it contains 60\% black pixels. The top-to-bottom scan terminates when a row that contains less than the threshold percentage of pixels is encountered. The same scanning procedure is repeated for the other three main directions. Fig.\ \ref{black_frame} shows the result of this procedure on a sample image.

\subsection{Approximate Lesion Localization}
\label{lesion_localization}
Although dermoscopy images can be quite large, the actual lesion often occupies a relatively small area. Therefore, if we can determine the approximate location of the lesion, the border detection algorithm can focus on this region rather than the whole image. An accurate bounding box (the smallest axis-aligned rectangular box that encloses the lesion) might be useful for various reasons:
\begin{inparaenum}[(i)]
\item it provides an estimate of the lesion size (certain image segmentation algorithms such as region growing and morphological flooding can use the size of the region as a termination criterion),
\item it might improve the border detection accuracy since the procedure is focused on a region that is guaranteed to contain the lesion,
\item it speeds up the border detection since the procedure is performed on a region that is often smaller than the whole image,
\item its surrounding might be utilized in the estimation of the background skin color, which is useful for various operations including the elimination of spurious regions that are discovered during the border detection procedure \cite{Celebi08a} and the extraction of dermoscopic features such as blotches \cite{Stoecker05} and blue-white areas \cite{Celebi08b}.
\end{inparaenum}
\par
\begin{figure}[!ht]
\centering
 \subfigure[Original image]{\label{threshold_a}\includegraphics[width=0.32\columnwidth]{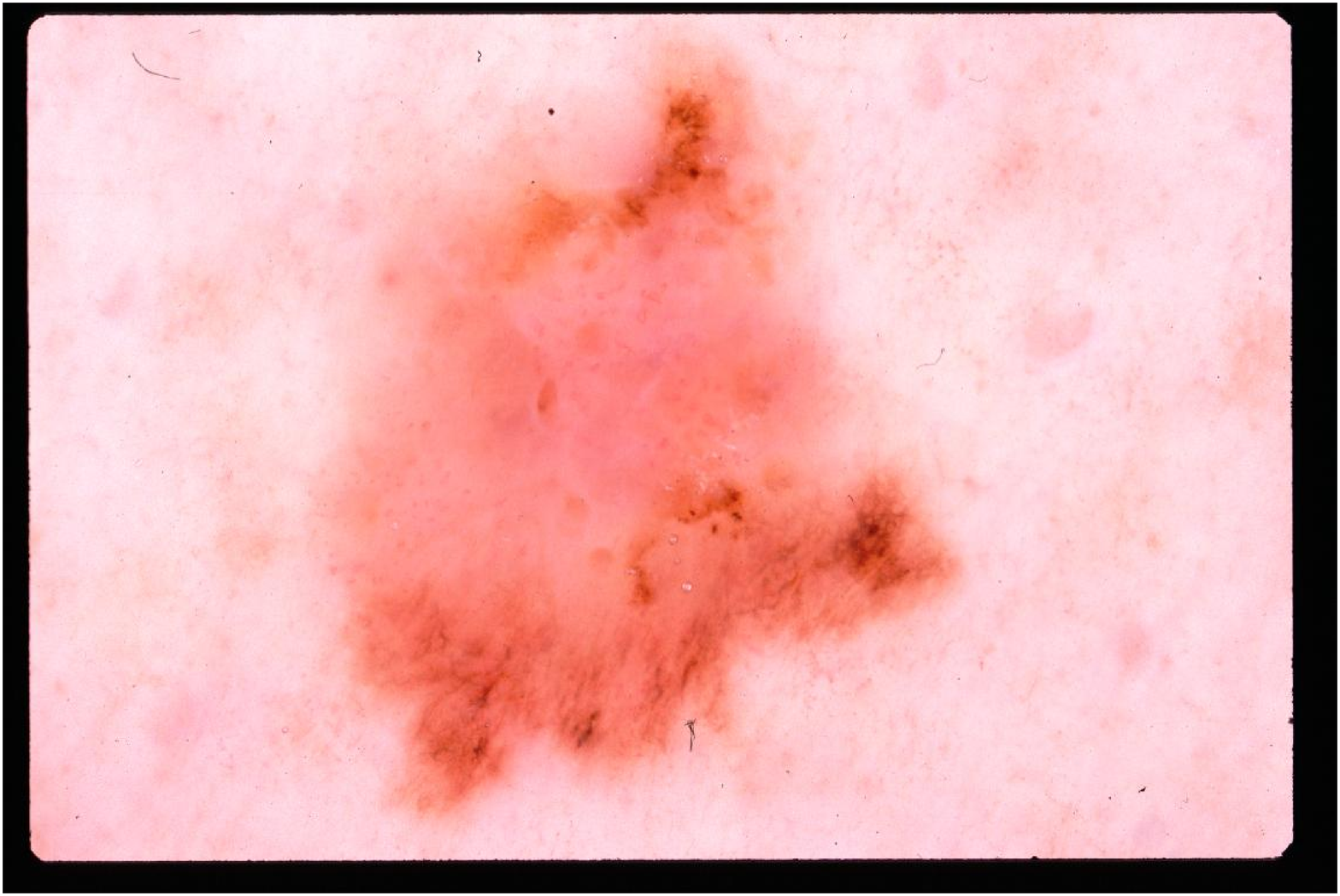}}
 \hspace{.2in}
 \subfigure[Blue channel]{\label{threshold_b}\includegraphics[width=0.32\columnwidth]{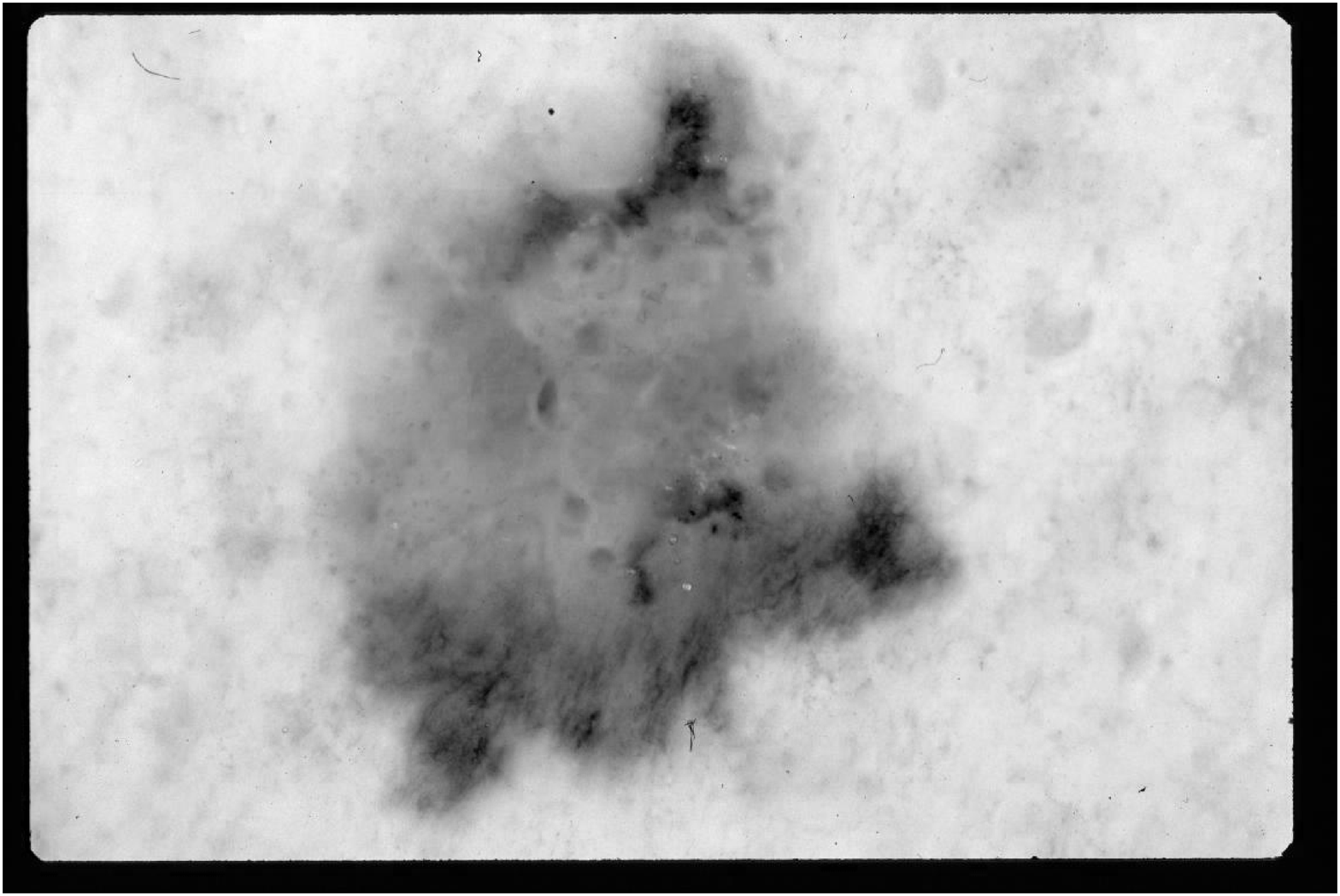}}
 \hspace{.2in}
 \subfigure[Otsu's method \cite{Otsu79} (\emph{T} = 137)]{\label{threshold_c}\includegraphics[width=0.32\columnwidth]{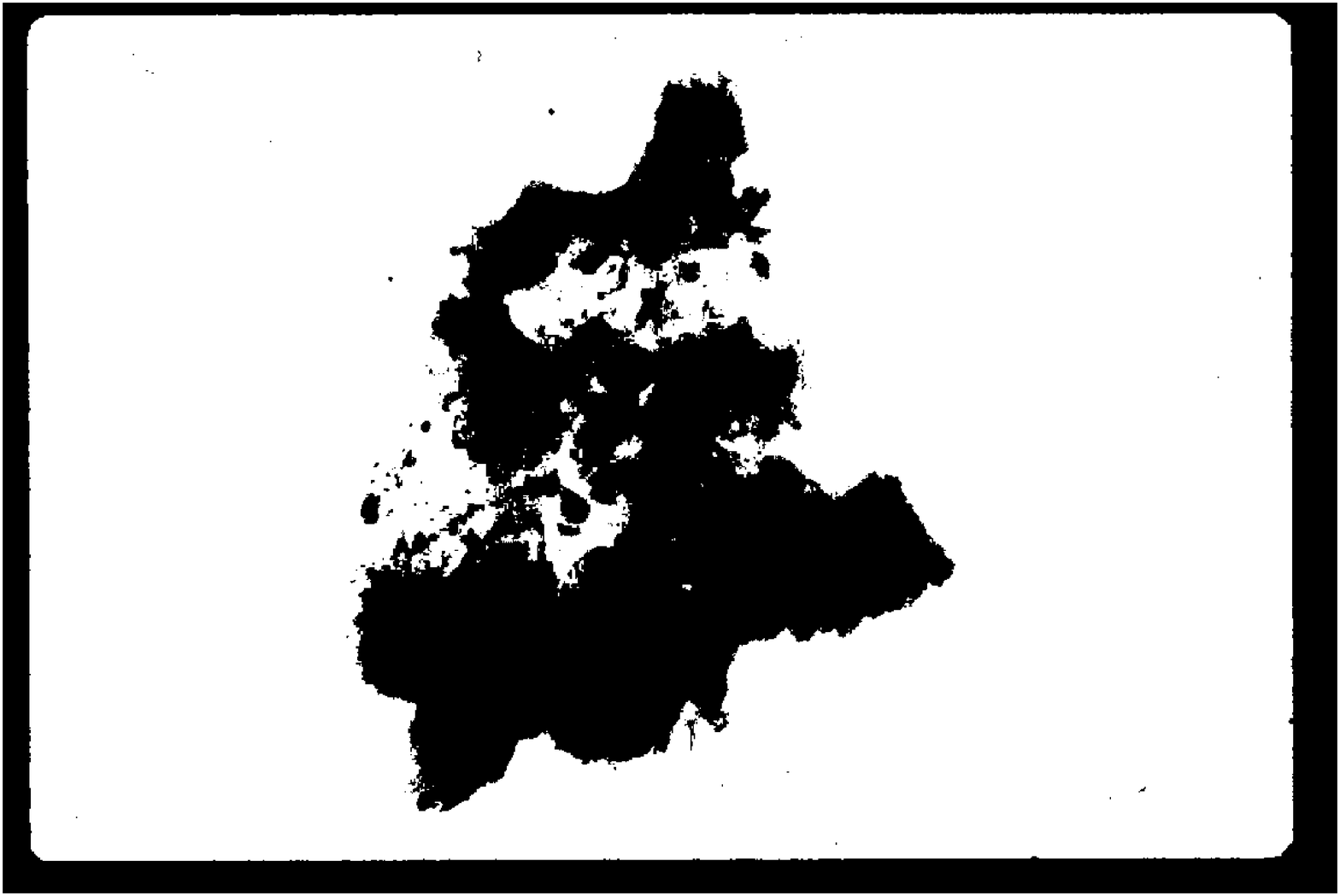}}
 \hspace{.2in}
 \subfigure[Kapur \emph{et al.}'s method \cite{Kapur85} (\emph{T} = 178)]{\label{threshold_d}\includegraphics[width=0.32\columnwidth]{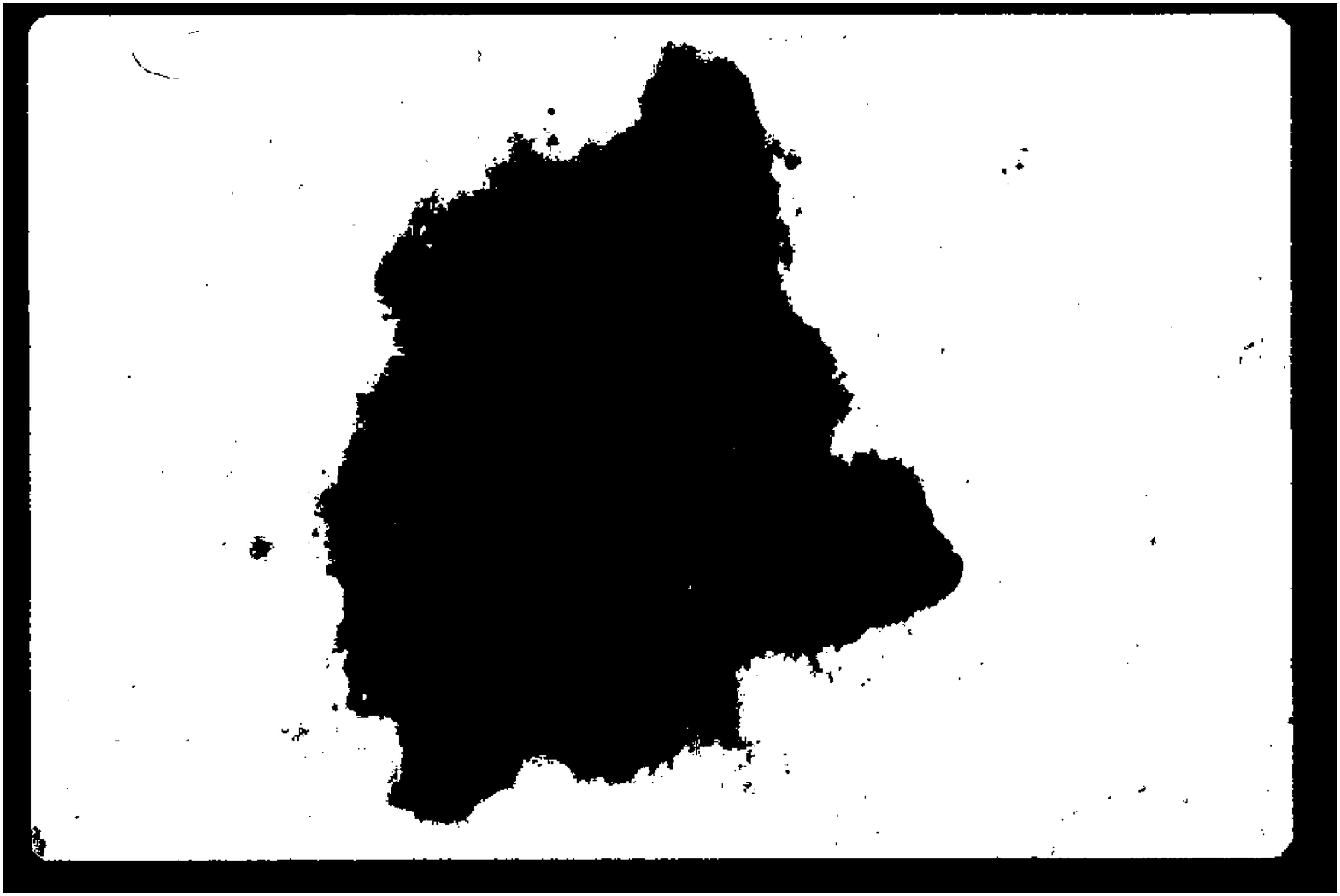}}
 \hspace{.2in}
 \subfigure[Huang \& Wang's method \cite{Huang95} (\emph{T} = 183)]{\label{threshold_e}\includegraphics[width=0.32\columnwidth]{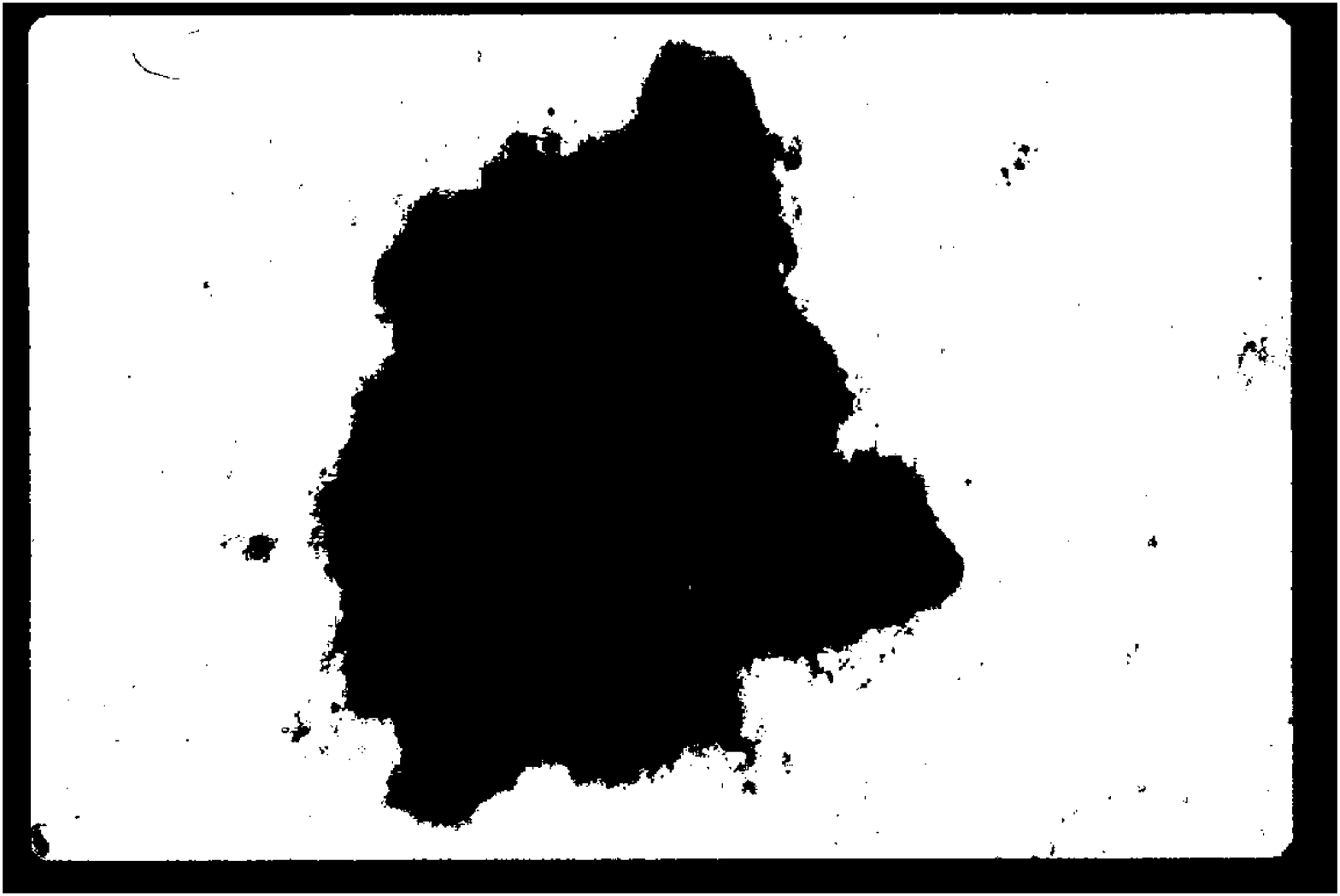}}
 \hspace{.2in}
 \subfigure[Yen \emph{et al.}'s method \cite{Yen95} (\emph{T} = 200)]{\label{threshold_f}\includegraphics[width=0.32\columnwidth]{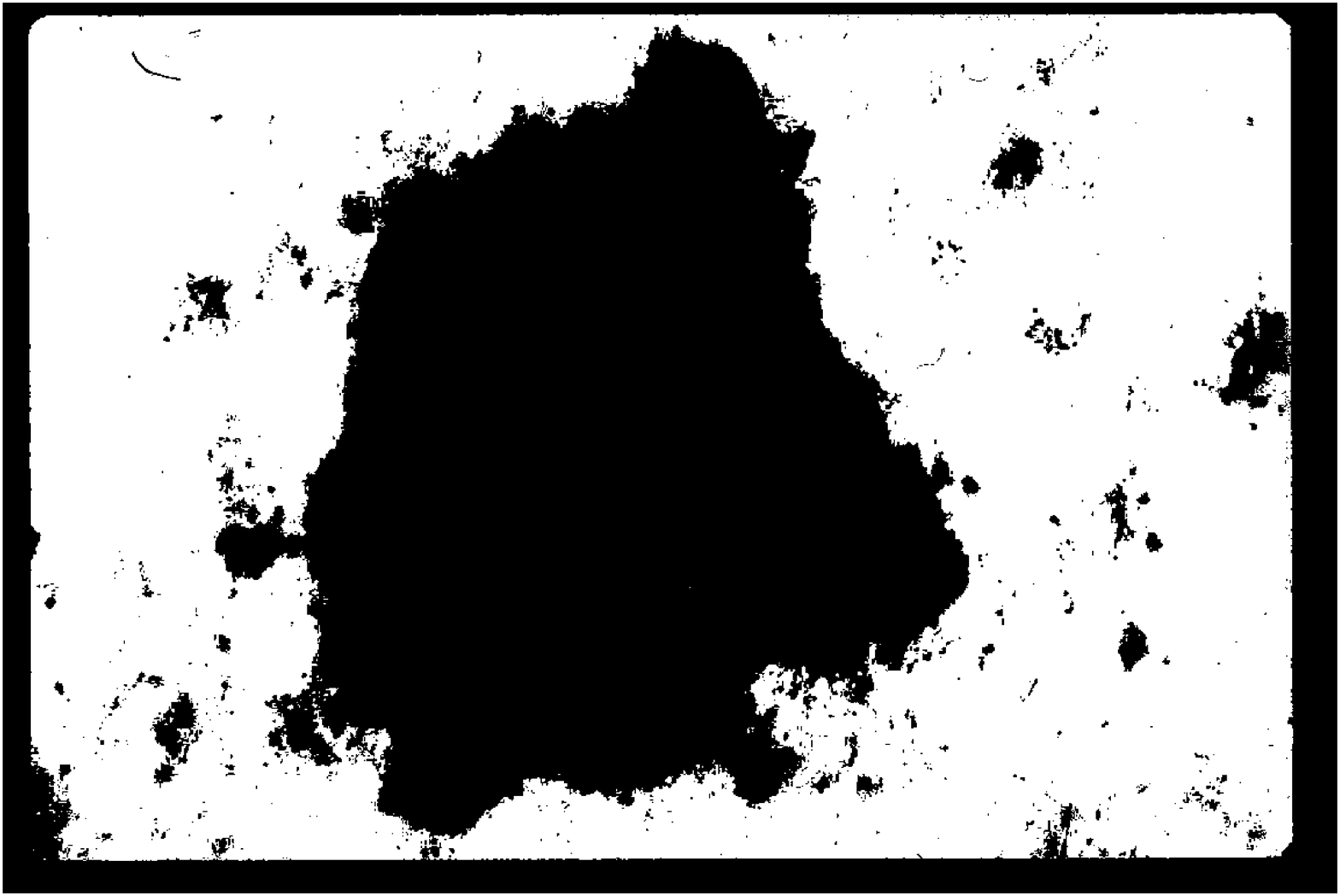}}
 \hspace{.2in}
 \subfigure[Sahoo \emph{et al.}'s method \cite{Sahoo97} (\emph{T} = 179)]{\label{threshold_g}\includegraphics[width=0.32\columnwidth]{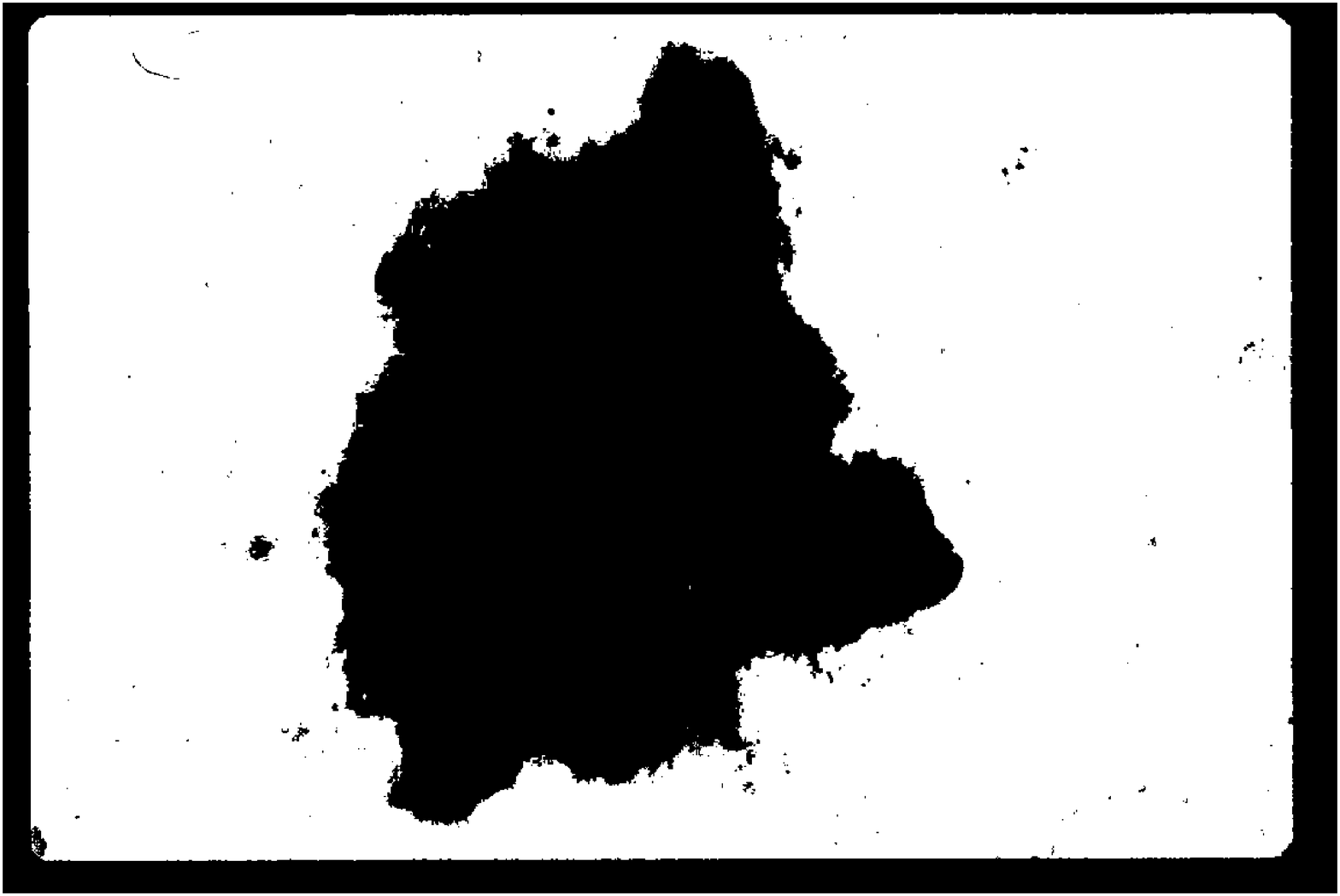}}
 \hspace{.2in}
 \subfigure[Li \& Tam's method \cite{Li98} (\emph{T} = 59)]{\label{threshold_h}\includegraphics[width=0.32\columnwidth]{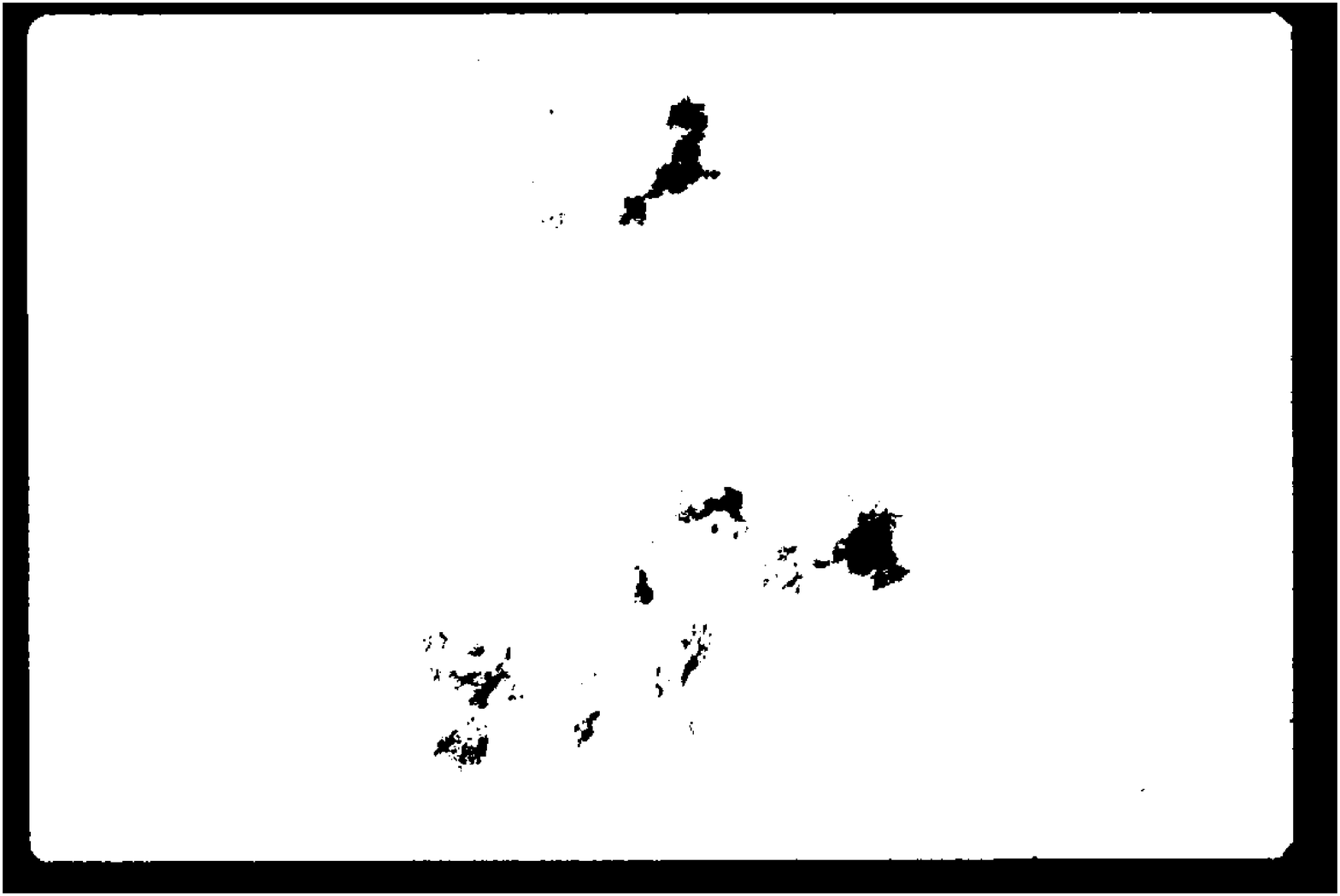}}
 \caption{Comparison of various thresholding methods (\emph{T}: threshold)}
 \label{threshold_comparison}
\end{figure}

In many dermoscopic images, the lesion can be roughly separated from the background skin using a grayscale thresholding method applied to the blue channel \cite{Iyatomi06,Celebi07}. While there are a number of thresholding methods that perform well in general, the effectiveness of a method strongly depends on the statistical characteristics of the image \cite{Melgani06}. Fig.\ \ref{threshold_comparison} illustrates this phenomenon\footnote{The frame of this image is left intact for visualization purposes.}. Here, methods \ref{threshold_d}, \ref{threshold_e}, and \ref{threshold_g} perform quite well. In contrast, methods \ref{threshold_c} and \ref{threshold_h} underestimate the optimal threshold, whereas method \ref{threshold_f} overestimates the optimal threshold. Although method \ref{threshold_c} is the most popular thresholding algorithm in the literature \cite{Sezgin04}, for this particular image, it performs the second worst.

A possible approach to overcome this problem is to fuse the results provided by an ensemble of thresholding algorithms. In this way, it is  possible to exploit the peculiarities of the participating thresholding algorithms synergistically, thus arriving at more robust final decisions than is possible with a single thresholding algorithm. We note that the goal of the fusion is not to outperform the individual thresholding algorithms, but to obtain accuracies comparable to that of the best thresholding algorithm independently of the image characteristics. In this study, we used the threshold fusion method proposed by Melgani \cite{Melgani06}, which we describe briefly in the  following.
\par
Let $X=\left\{x_{mn}: m=0,1,\ldots,M-1,\ n=0,1,\ldots,N-1\right\}$ be the original scalar $M \times N$ image with $L$ possible gray levels $\left(x_{mn}\in\left\{0,1,\ldots,L-1\right\}\right)$ and $Y=\left\{y_{mn}: m=0,1,\ldots,M-1,\ n=0,1,\ldots,N-1\right\}$ be the binary output of the threshold fusion. Consider an ensemble of $P$ thresholding algorithms. Let $T_i$ and $A_i \ (i=1,2,\ldots,P)$ be the threshold value and the output binary image associated with the i-th algorithm of the ensemble, respectively. Within a Markov Random Field (MRF) framework the fusion problem can be formulated as an energy minimization task. Accordingly, the local energy function $\U_{mn}$ to be minimized for the pixel $(m,n)$ can be written as follows:
\begin{equation}
\label{energy_function}
\U_{mn} = \beta_{SP} \cdot \U_{SP}\left[y_{mn},Y^S(m,n)\right] + \sum_{i=1}^P { \beta_i \cdot \U_{II}\left[y_{mn},A^S_{i}(m,n)\right] }
\end{equation}

where $S$ is a predefined neighborhood system associated with pixel $(m,n)$, $\U_{SP}(\cdot)$ and $\U_{II}(\cdot)$ refer  to the spatial and inter-image energy functions, respectively, whereas $\beta_{SP}$ and $\beta_i \ (i=1,2,\ldots,P)$ represent the spatial and inter-image parameters, respectively. The spatial energy function can be expressed as:
\begin{equation}
\U_{SP}\left[y_{mn},Y^S(m,n)\right] = - \sum_{ y_{pq} \in Y^S(m,n) } \I\left(y_{mn},y_{pq}\right)
\end{equation}

where $\I(.,.)$ is the indicator function defined as:
\begin{equation}
\I{\left(y_{mn},y_{pq}\right)}= \left\{
                 \begin{array}
                 {r@{\quad}l}
                 1 & \mbox{if} \ y_{mn}=y_{pq} \\
                 0 & \mbox{otherwise}
                 \end{array}
          \right.
\end{equation}

The inter-image energy function is defined as:
\begin{equation}
\U_{II}\left[y_{mn},A^S(m,n)\right] = - \sum_{ A_i(p,q) \in A^S_{i}(m,n) } \alpha^i(x_{pq}) \cdot \I\left[y_{mn},A_i(p,q)\right]
\end{equation}

where $\alpha^i(\cdot)$ is a weight function given by:
\begin{equation}
\label{alpha}
\alpha^i(x_{mn}) = 1 - \exp \left( -\gamma \left| x_{mn} - T_i \right| \right)
\end{equation}

This function controls the effect of unreliable decisions at the pixel level that can be incurred by the thresholding algorithms. At the global (image) level decisions are weighed by the inter-image parameters $\beta_i \ (i=1,2,\ldots,P)$, which are computed as follows:
\begin{equation}
\label{beta}
\beta_i = \exp \left( -\gamma \left| \bar{T} - T_i \right| \right)
\end{equation}

where $\bar{T}$ is the average threshold value:
\begin{equation}
\bar{T} = \frac {1}{P} \sum_{i=1}^P T_i
\end{equation}

The MRF fusion strategy proposed in \cite{Melgani06} is as follows:

\begin{enumerate}
	\item Apply each thresholding algorithm of the ensemble to the image $X$ to generate the set of thresholded images $A_i \ (i=1,2,\ldots,P)$
	\item Initialize $Y$ by minimizing for each pixel $(m,n)$ the local energy function $\U_{mn}$ defined in Eq.\ \ref{energy_function} without the spatial energy term i.e., by setting $\beta_{SP}=0$.
	\item Update $Y$ by minimizing for each pixel $(m,n)$ the local energy function $\U_{mn}$ defined in Eq.\ \ref{energy_function} including the spatial energy term i.e., by setting $\beta_{SP} \neq 0$.
	\item Repeat step 3 $K_{max}$ times or until the number of different labels in $Y$ computed over the last two iterations becomes very small.
\end{enumerate}

In our preliminary experiments, we observed that, besides being computationally demanding, the iterative part (step 3) of the fusion algorithm makes only marginal contribution to the quality of the results. Therefore, in this study, we considered only the first two steps. The $\gamma$ parameter was set to the recommended value of $0.1$ \cite{Melgani06}. For computational reasons, $\alpha$ (Eq. \ref{alpha}) and $\beta$ (Eq. \ref{beta}) values were precalculated and the neighborhood system $S$ was chosen as a $3 \times 3$ square.
\par
The most important performance factor in the fusion algorithm seems to be the choice of the thresholding algorithms. We considered six popular thresholding algorithms to construct the ensemble: Otsu's \cite{Otsu79}, Kapur \emph{et al.}'s \cite{Kapur85}, Huang \& Wang's \cite{Huang95}, Yen \emph{et al.}'s \cite{Yen95} , Sahoo \emph{et al.}'s \cite{Sahoo97}, and Li \& Tam's \cite{Li98} methods. In order to determine the best combination, we evaluated ensembles with 3 (20 ensembles), 4 (15 ensembles), 5 (6 ensembles), and 6 (1 ensembles) methods.
\par
\begin{figure}[!ht]
\centering
 \subfigure[Otsu-Kapur-Huang]{\label{ensemble_a}\includegraphics[width=0.4\columnwidth]{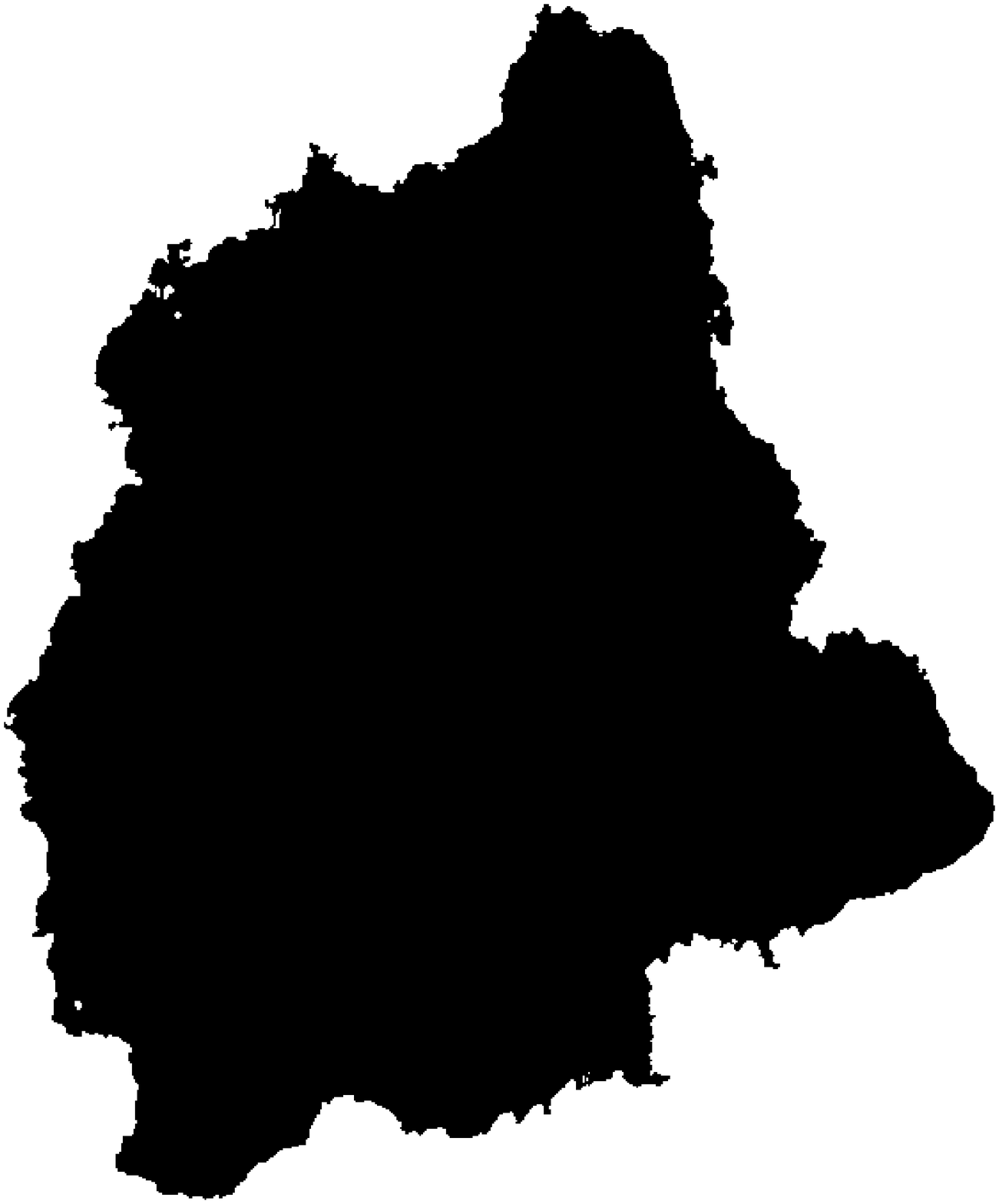}}
 \subfigure[Yen-Sahoo-Li]{\label{ensemble_b}\includegraphics[width=0.4\columnwidth]{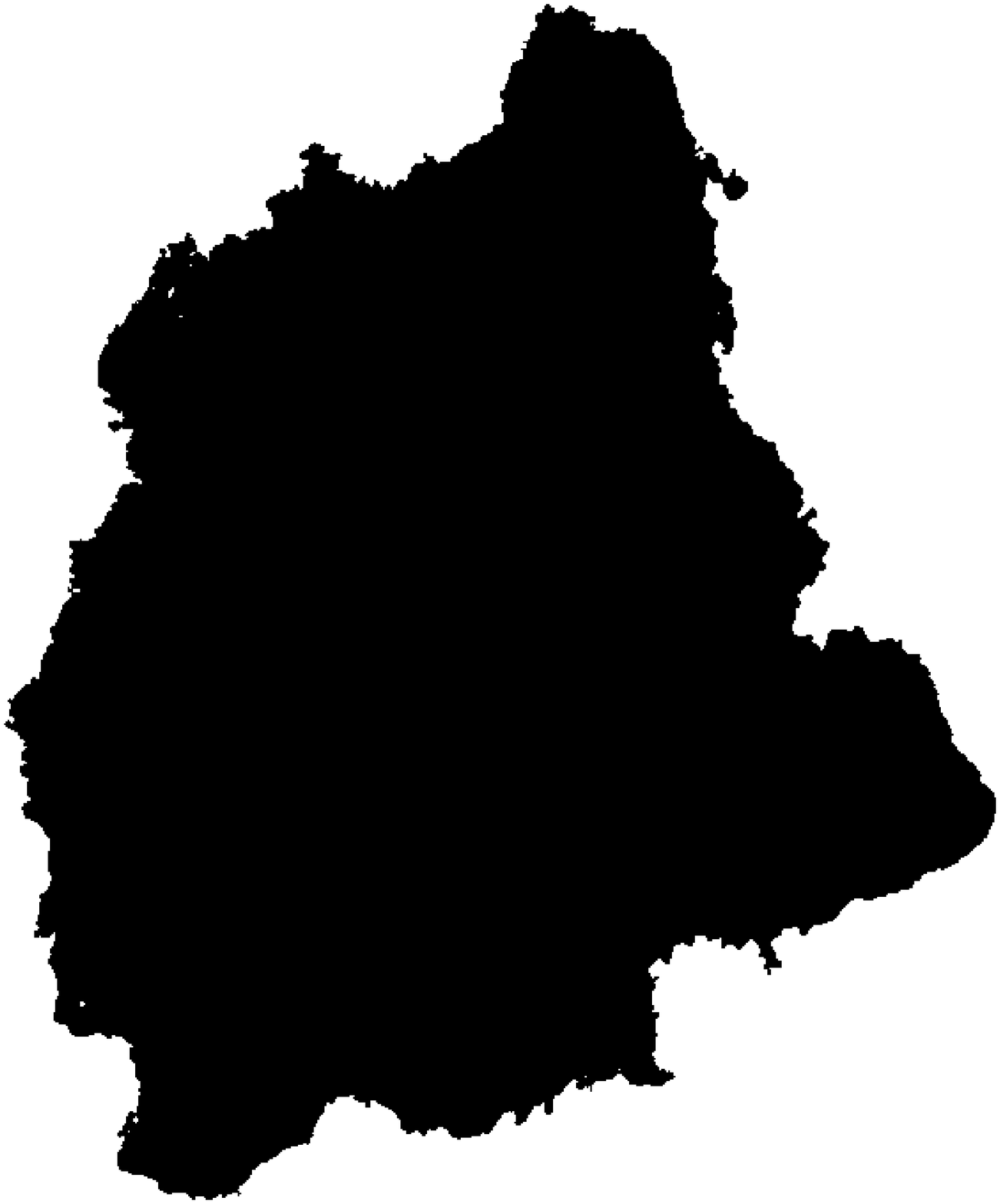}}
 \subfigure[Otsu-Kapur-Huang-Yen]{\label{ensemble_c}\includegraphics[width=0.4\columnwidth]{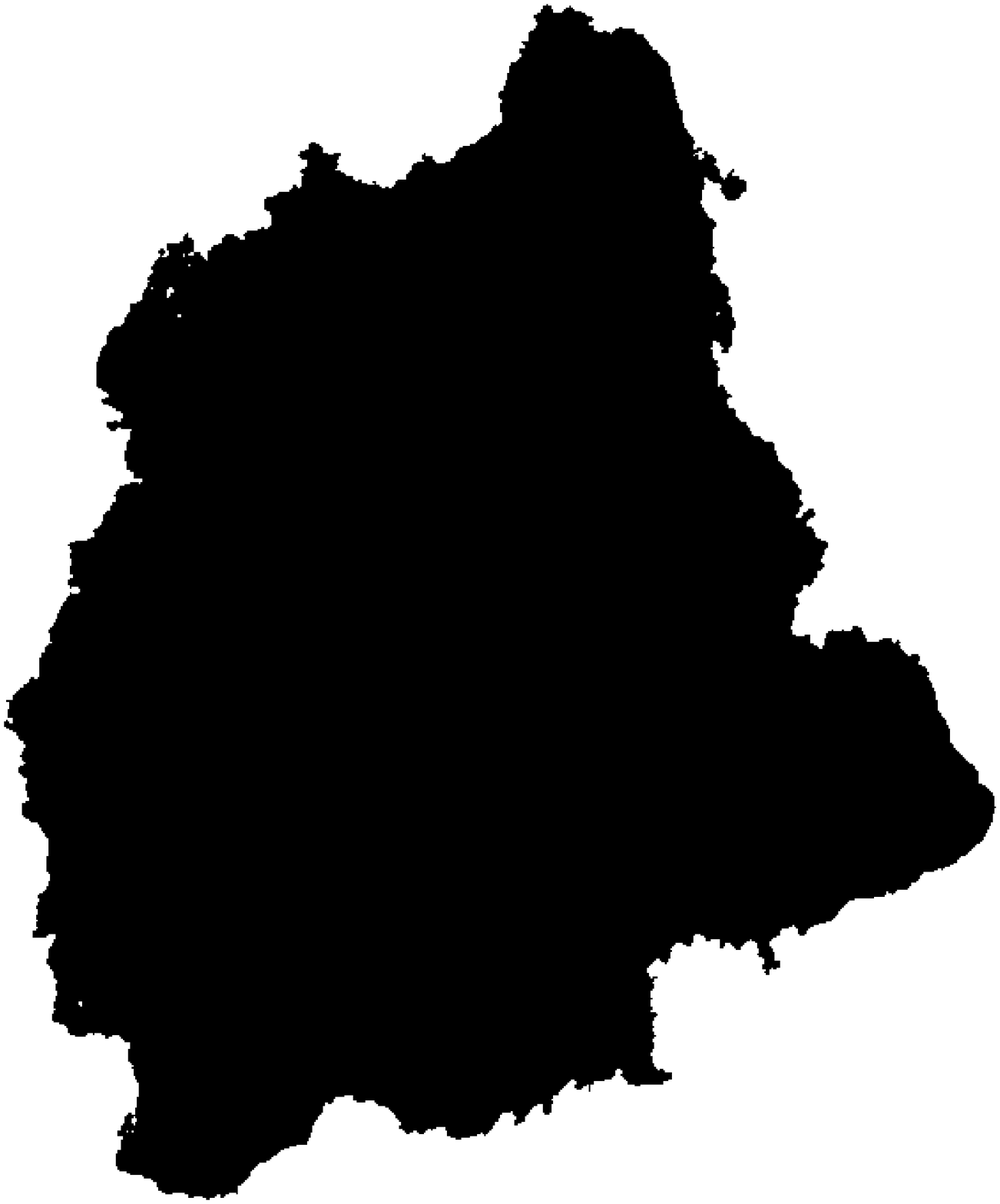}}
 \subfigure[Huang-Yen-Sahoo-Li]{\label{ensemble_d}\includegraphics[width=0.4\columnwidth]{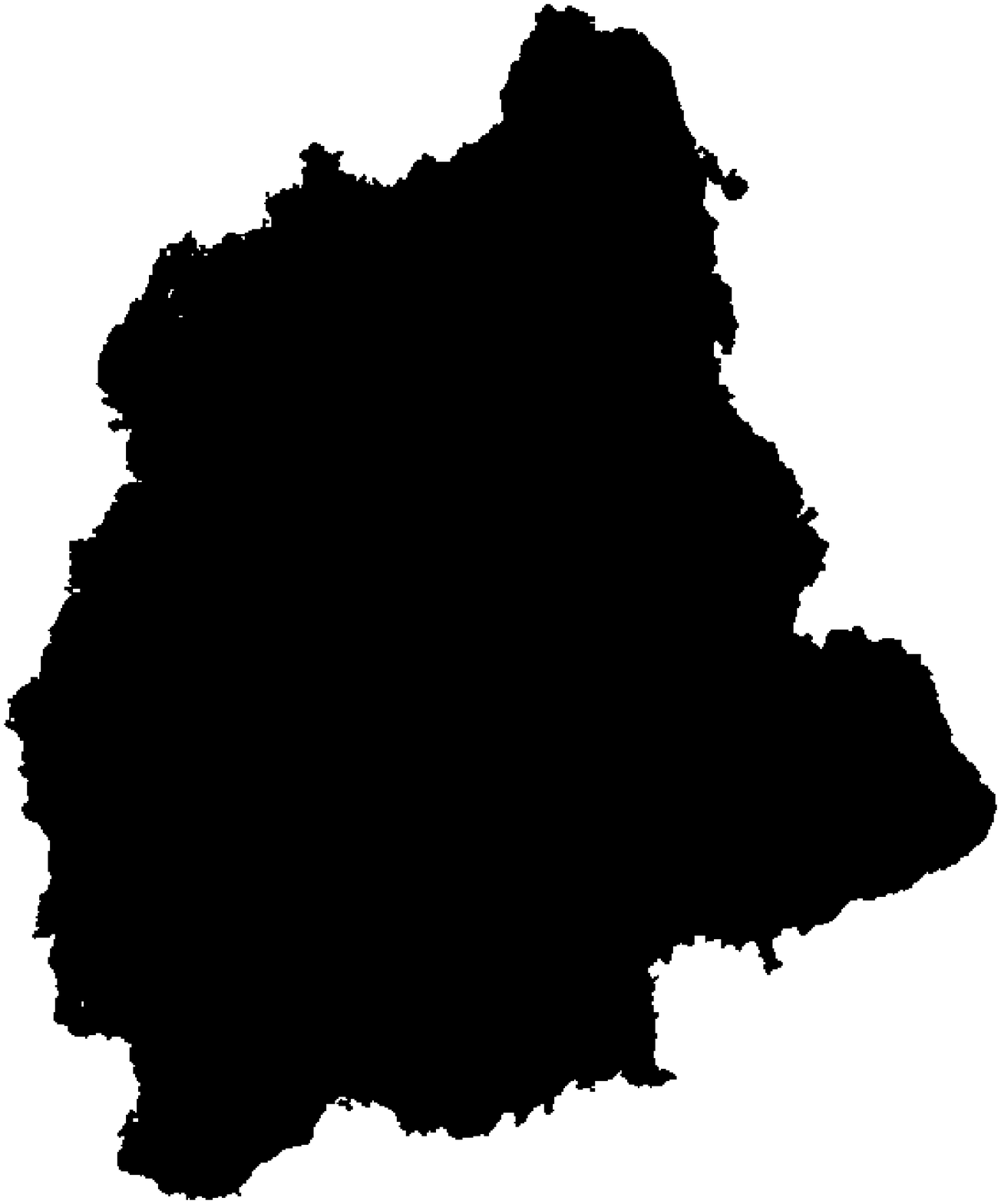}}
 \caption{Comparison of various threshold ensembles}
 \label{ensemble_comparison}
\end{figure}

Fig.\ \ref{ensemble_comparison} shows the output of four particular ensembles: Otsu-Kapur-Huang, Yen-Sahoo-Li, Otsu-Kapur-Huang-Yen, and Huang-Yen-Sahoo-Li. Note that each ensemble contains at least one method that either underestimates or overestimates the optimal threshold. It can be seen that each ensemble performs equally well, which demonstrates that failures in pathological cases might be prevented using a proper fusion strategy.

\begin{figure}[!ht]
\centering
 \subfigure[Non-adaptive expansion with $P = 3\%$]{\label{expansion_a}\includegraphics[width=0.4\columnwidth]{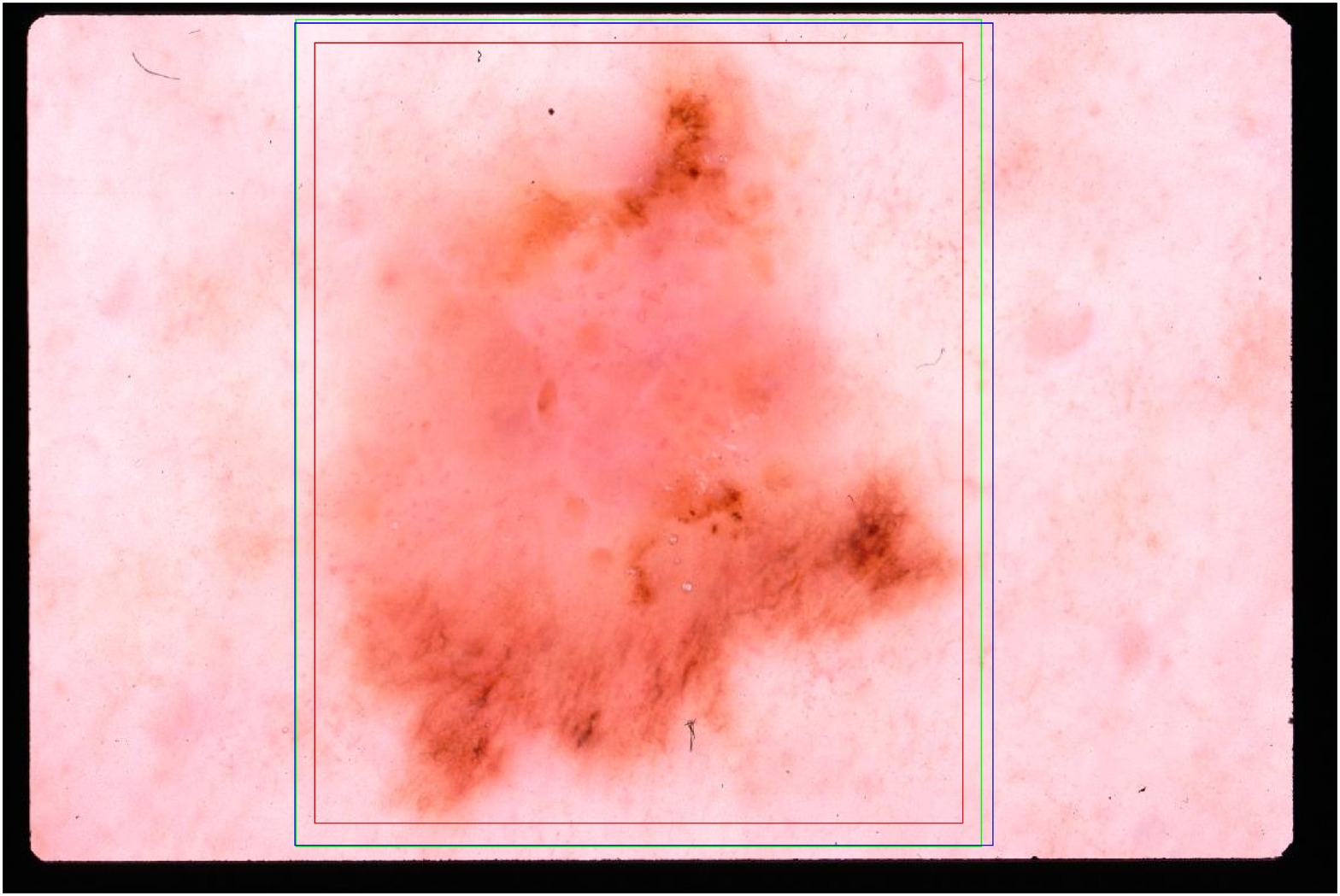}}
 \hspace{.2in}
 \subfigure[Adaptive expansion with $G = 6$]{\label{expansion_b}\includegraphics[width=0.4\columnwidth]{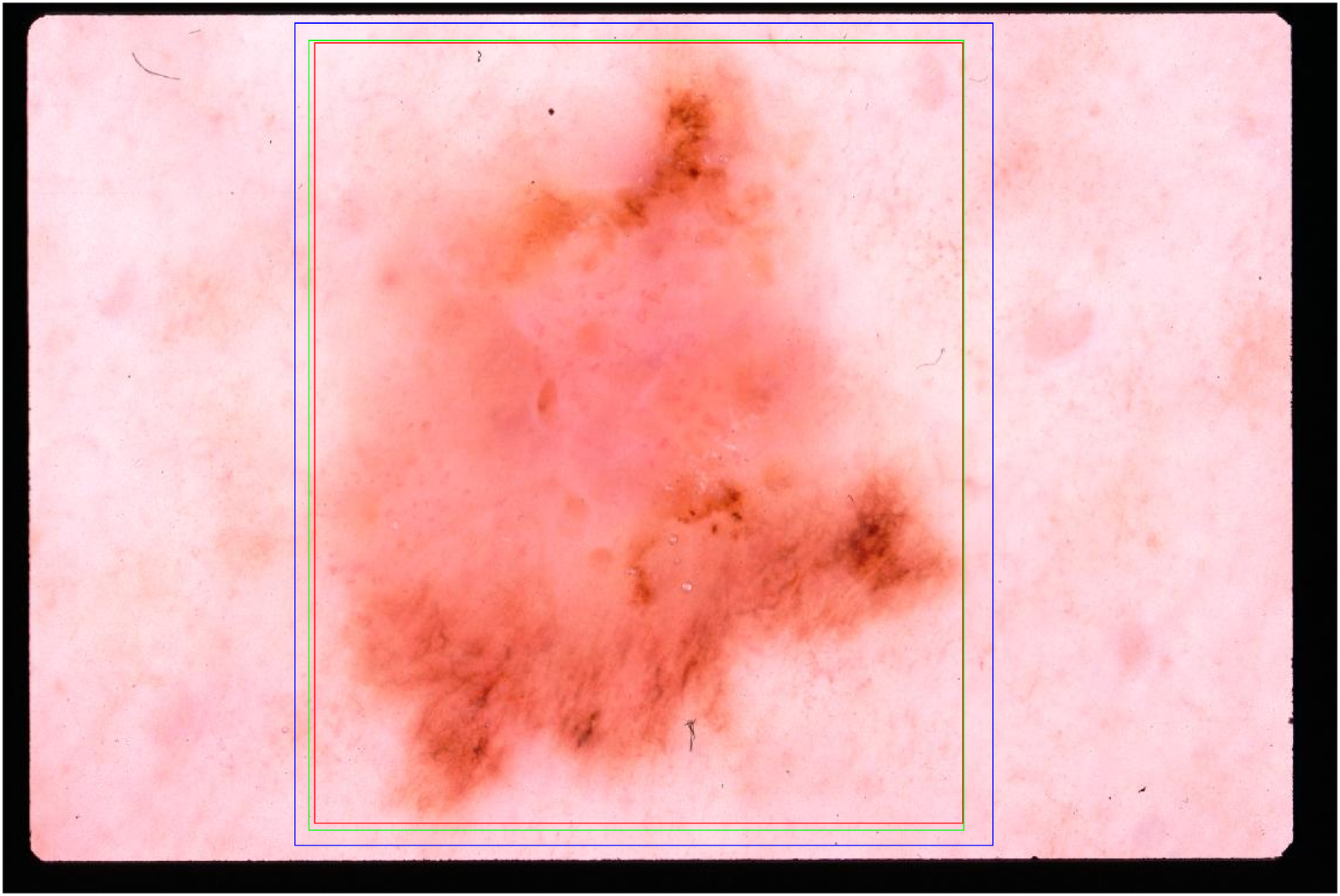}}
 \caption{Comparison of the bounding box expansion methods}
 \label{expansion_comparison}
\end{figure}

Fig.\ \ref{expansion_a} shows the result of the ensemble Otsu-Kapur-Huang-Sahoo. Here, the blue bounding box encloses the dermatologist determined border (see Section \ref{results_and_discussion}), whereas the red one encloses the binary output of the threshold fusion. It can be seen that the red box is completely contained within the blue box. This was observed in many cases because the automated thresholding methods tend to find the sharpest pigment change, whereas the dermatologists choose the outmost detectable pigment. We experimented with two different expansion methods to solve this problem. The first one involves expanding the automatic box by $P$\% in four main directions. In other words, an automatic box of size $M_B \times N_B$ is expanded by $M_B \cdot P/100$ pixels in the West and East directions and $N_B \cdot P/100$ pixels in the North and South directions. The second one involves incrementing the threshold values obtained by each algorithm in the ensemble by $G$ gray levels. In the rest of this article, we will refer to these expansion methods as non-adaptive and adaptive, respectively. Figs.\ \ref{expansion_a} and \ref{expansion_b} show the results of these methods with the expanded box shown in green. In this particular example, the non-adaptive method performs better in bringing the automatic box closer to the manual one. In order to determine the optimal expansion amounts we evaluated $P \in \left\{2, 4, 6, 8\right\}$ and $G \in \left\{4, 6, 8, 10\right\}$.

\section{Results and Discussion}
\label{results_and_discussion}
The proposed method was tested on a set of 428 dermoscopy images obtained from the EDRA Interactive Atlas of Dermoscopy \cite{Argenziano02} and the Keio University Hospital. These were 24-bit RGB color images with dimensions ranging from $771 \times 507$ pixels to $768 \times 512$ pixels. An experienced dermatologist (WVS) determined the manual borders by selecting a number of points on the lesion border, which were then connected by a second-order B-spline. The bounding box error was quantified using the grading system developed by Hance \emph{et al.} \cite{Hance96}\:
\begin{equation}
\mbox{\emph{$\varepsilon$}} = \frac { \mbox{Area(\emph{AutomaticBox}} \oplus \mbox{\emph{ManualBox})} }
                                    { \mbox{Area(\emph{ManualBox})} } \cdot 100
\end{equation}

where \emph{AutomaticBox} is the binary image obtained by filling the bounding box of the fusion output, \emph{ManualBox} is the binary image  obtained by filling the bounding box of the dermatologist-determined border, $\oplus$ is the exclusive-OR operation, which essentially determines the pixels for which the \emph{AutomaticBox} and \emph{ManualBox} disagree, and $\mbox{Area}(I)$ denotes the number of pixels in the binary image $I$.
\par
We determined the optimal parameter combination for the presented approximate bounding box computation method as follows. First, the black frame removal procedure described in Section \ref{frame_removal} is performed on each image in the data set. The lesion bounding box is then computed using the fusion method described in Section \ref{lesion_localization} with one of the 42 ensembles. Finally, the approximate bounding box is expanded using either the non-adaptive method with $P \in \left\{2, 4, 6, 8\right\}$ or the adaptive method with $G \in \left\{4, 6, 8, 10\right\}$. Table \ref{ensemble_table} shows various statistics associated with the four most accurate ensembles for each expansion method. The last two columns refer to the mean and standard deviation values, respectively for the percentage image size reduction, i.e.\ $\frac { \mbox{Area(\emph{AutomaticBox})}} {M \cdot N} \cdot 100$, provided by the bounding box computation. The following observations are in order:
\begin{inparaenum}[(i)]
\item both expansion methods reduce the mean bounding box error,
\item the lowest mean errors were obtained using the ensemble Otsu-Kapur-Huang-Sahoo,
\item the non-adaptive expansion method was more effective than the adaptive one,
\item the computation of the bounding box reduced the original image size by about 260\%.
\end{inparaenum}

\begin{table}
\centering
\caption{ \label{ensemble_table} Ensemble statistics ($\mu$: mean, $\sigma$: std.\ dev., $\varepsilon_i$: initial box error, $\varepsilon_x$: expanded box error)}
\begin{tabular}{|c|c|c|c|c|c|c|c|}
\hline
\textbf{Ensemble} & \textbf{Expansion Method} & $\mu_{\varepsilon_i}$ & $\sigma_{\varepsilon_i}$ & $\mu_{\varepsilon_x}$ & $\sigma_{\varepsilon_x}$ & $\mu_s$ & $\sigma_s$\\
\hline
\hline
Otsu-Kapur-Huang-Sahoo  & Non-adaptive ($P=2$) & 10.25 & 8.10 & 7.58 & 8.13 & 268.31 & 185.64\\
\hline
Otsu-Huang-Yen-Li  & Non-adaptive ($P=4$) & 11.92 & 7.59 & 7.89 & 6.30 & 260.55 & 183.85\\
\hline
Otsu-Huang-Sahoo-Li  & Non-adaptive ($P=4$) & 11.98 & 7.62 & 7.90 & 6.20 & 260.95 & 184.14\\
\hline
Otsu-Huang-Sahoo  & Non-adaptive ($P=2$) & 11.14 & 7.17 & 7.91 & 6.71 & 273.84 & 195.69\\
\hline
Otsu-Kapur-Huang-Sahoo  & Adaptive ($G=6$) & 10.25 & 8.10 & 9.27 & 7.68 & 276.92 & 192.14\\
\hline
Kapur-Huang-Sahoo-Li  & Adaptive ($G=8$) & 10.98 & 7.66 & 9.43 & 7.69 & 279.03 & 194.42\\
\hline
Otsu-Kapur-Huang-Sahoo  & Adaptive ($G=4)$ & 10.25 & 8.10 & 9.44 & 7.56 & 279.98 & 194.26\\
\hline
Kapur-Huang-Sahoo-Li  & Adaptive ($G=6$) & 10.98 & 7.66 & 9.67 & 7.58 & 282.09 & 196.58\\
\hline
\end{tabular}
\end{table}

The adaptive method was less effective than the non-adaptive one probably because the former often expands the approximate box by unpredictable amounts: either too little (as in Fig.\ \ref{expansion_b}) or too much depending on the shape of the histogram and the value of the $G$ parameter. In contrast, the latter always expands the approximate box by an amount specified by the $P$ parameter.
\par
Table \ref{individual_table} shows the statistics for the individual thresholding methods. Note that, due to space limitations, we report only the results of the non-adaptive expansion method (as in the ensemble case, the adaptive method has inferior performance). It can be seen that, in most configurations, the individual methods obtain significantly higher mean errors than the best ensemble methods, i.e.\ the first four rows of Table \ref{ensemble_table}. This is because, as explained in Section \ref{lesion_localization}, the individual methods are more prone to catastrophic failures when given pathological input images. The high standard deviation values also support this explanation. Only the performance of Otsu (with $P = 2,4$) and Li \emph{et al.}'s (with $P = 4$) methods is close to the performance of the ensembles. However, as mentioned in Section \ref{lesion_localization}, the goal of fusion is not to outperform the individual thresholding algorithms, but to obtain accuracies comparable to that of the best thresholding algorithm independently of the image characteristics.

\begin{table}
\centering
\caption{ \label{individual_table} Individual statistics ($\mu$: mean, $\sigma$: std.\ dev., $\varepsilon_i$: initial box error, $\varepsilon_x$: expanded box error)}
\begin{tabular}{|c|c|c|c|c|c|c|c|}
\hline
\textbf{Thresholding Method} & \textbf{Expansion Method} & $\mu_{\varepsilon_i}$ & $\sigma_{\varepsilon_i}$ & $\mu_{\varepsilon_x}$ & $\sigma_{\varepsilon_x}$ & $\mu_s$ & $\sigma_s$\\
\hline
\hline
Otsu  & Non-adaptive ($P=2$) & 12.05 & 9.10 & 9.00 & 8.95 & 275.07 & 199.28\\
\hline
Kapur  & Non-adaptive ($P=2$) & 12.87 & 16.86 & 12.68 & 17.56 & 261.95 & 197.94\\
\hline
Huang  & Non-adaptive ($P=2$) & 20.31 & 67.97 & 17.17 & 69.76 & 269.59 & 190.09\\
\hline
Yen  & Non-adaptive ($P=2$) & 14.98 & 27.12 & 15.74 & 27.74 & 255.61 & 250.53\\
\hline
Sahoo  & Non-adaptive ($P=2$) & 13.43 & 24.60 & 13.37 & 25.19 & 254.43 & 184.36\\
\hline
Li  & Non-adaptive ($P=2$) & 15.12 & 9.65 & 11.06 & 9.07 & 293.54 & 215.80\\
\hline
Otsu  & Non-adaptive ($P=4$) & 12.05 & 9.10 & 9.10 & 9.14 & 256.86 & 182.82\\
\hline
Kapur  & Non-adaptive ($P=4$) & 12.87 & 16.86 & 15.54 & 18.61 & 245.36 & 183.78\\
\hline
Huang  & Non-adaptive ($P=4$) & 20.31 & 67.97 & 16.83 & 70.69 & 251.99 & 174.44\\
\hline
Yen  & Non-adaptive ($P=4$) & 14.98 & 27.12 & 19.32 & 28.49 & 239.46 & 230.91\\
\hline
Sahoo  & Non-adaptive ($P=4$) & 13.43 & 24.60 & 16.43 & 25.98 & 238.32 & 170.38\\
\hline
Li  & Non-adaptive ($P=4$) & 15.12 & 9.65 & 9.41 & 7.99 & 273.93 & 198.61\\
\hline
\end{tabular}
\end{table}

As mentioned in Section \ref{lesion_localization}, an accurate bounding box can provide an estimate of the lesion size, i.e. $\mbox{Area(\emph{ManualBorder})}$. In order to verify this, we calculated the best fitting line for $\mbox{Area(\emph{AutomaticBox})}$ vs.\ $\mbox{Area(\emph{ManualBorder})}$ using the generalized least-squares method \cite{Li84}:
\begin{equation}
\label{auto_box_vs_man_border}
\mbox{Area(\emph{ManualBorder})} \approx \mbox{Area(\emph{AutomaticBox})} \cdot 0.861271 - 15627.226419
\end{equation}
where \emph{ManualBorder} is the binary image obtained by filling the dermatologist-determined border. The accuracy of this relation was calculated by plugging the area of the approximate bounding box for each image into Eq.\ \ref{auto_box_vs_man_border}, and then comparing the result with the actual area of the lesion, which is calculated from the dermatologist-determined border. The percentage mean and standard deviation errors over the entire image set were $11.88$ and $11.49$, respectively. These results demonstrate that the lesion size can be estimated from the bounding box area with relatively high accuracy. An even better estimate can be made from the binary output of the threshold fusion. The best fitting line for $\mbox{Area(\emph{FusionOutput})}$ vs.\ $\mbox{Area(\emph{ManualBorder})}$ was calculated as:
\begin{equation}
\label{fusion_output_vs_man_border}
\mbox{Area(\emph{ManualBorder})} \approx \mbox{Area(\emph{FusionOutput})} \cdot 1.158209 - 4485.287871
\end{equation}
where \emph{FusionOutput} is the binary output of the threshold fusion. The percentage mean and standard deviation errors for this relation were $8.16$ and $8.54$, respectively.
\par
\begin{figure}[!ht]
\centering
 \subfigure[$\varepsilon_i = 3.26\% \mbox{\; , \;} \varepsilon_x = 1.83 \%$]{\label{results_a}\includegraphics[width=0.4\columnwidth]{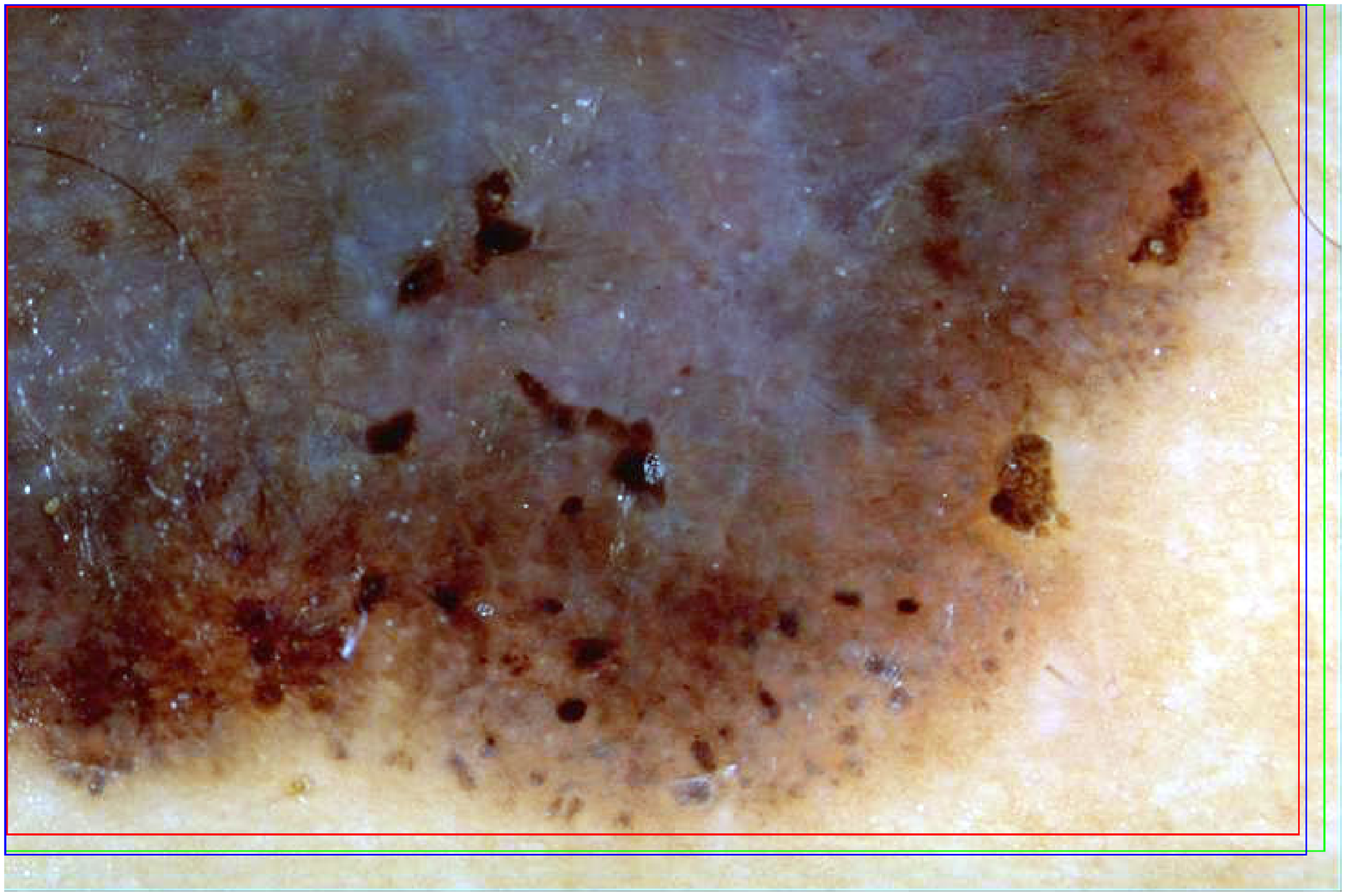}}
 \hspace{.2in}
 \subfigure[$\varepsilon_i = 4.89\% \mbox{\; , \;} \varepsilon_x = 3.62
\%$]{\label{results_b}\includegraphics[width=0.4\columnwidth]{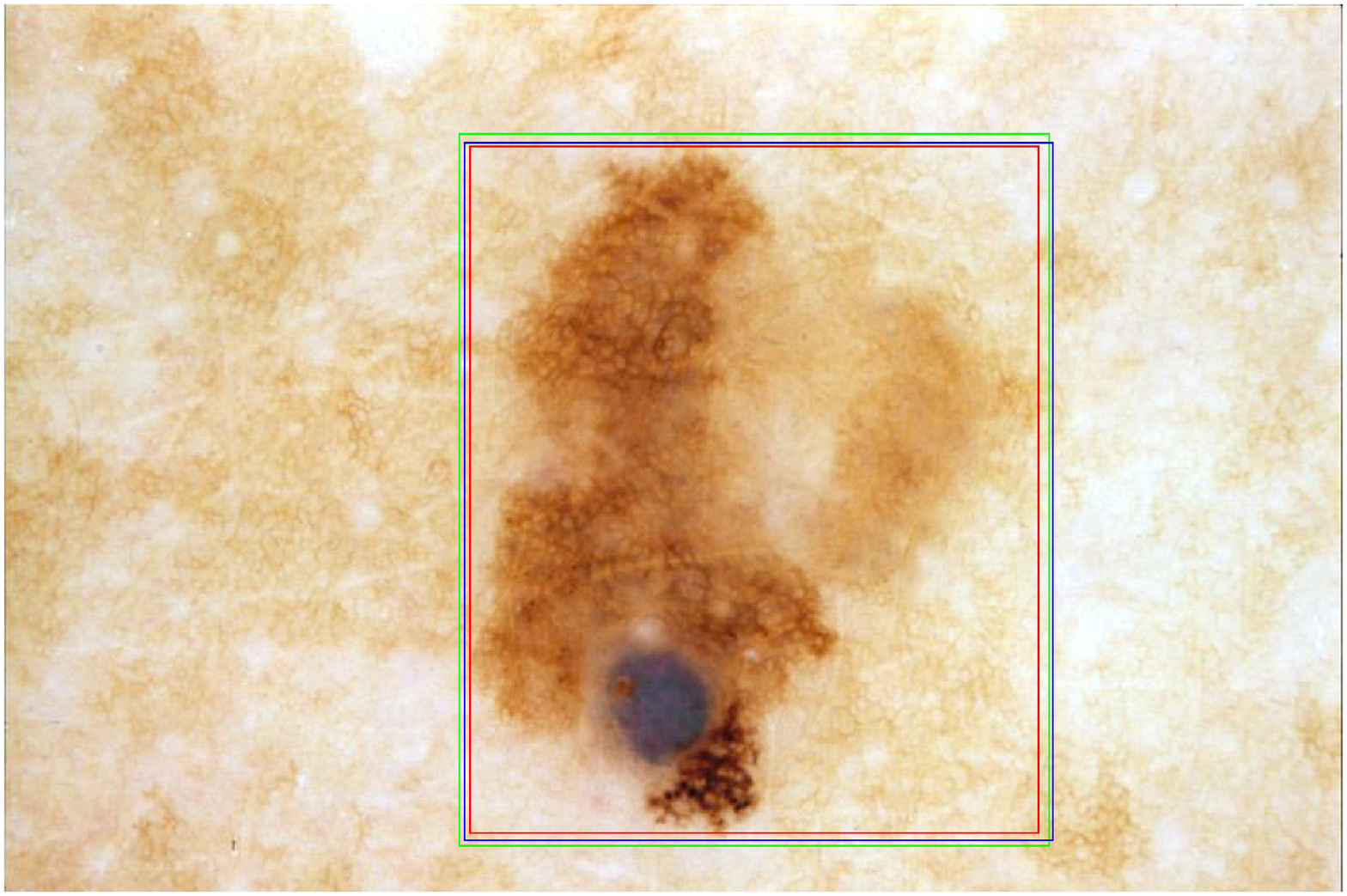}}
 \subfigure[$\varepsilon_i = 10.87\% \mbox{\; , \;} \varepsilon_x = 4.03
\%$]{\label{results_c}\includegraphics[width=0.4\columnwidth]{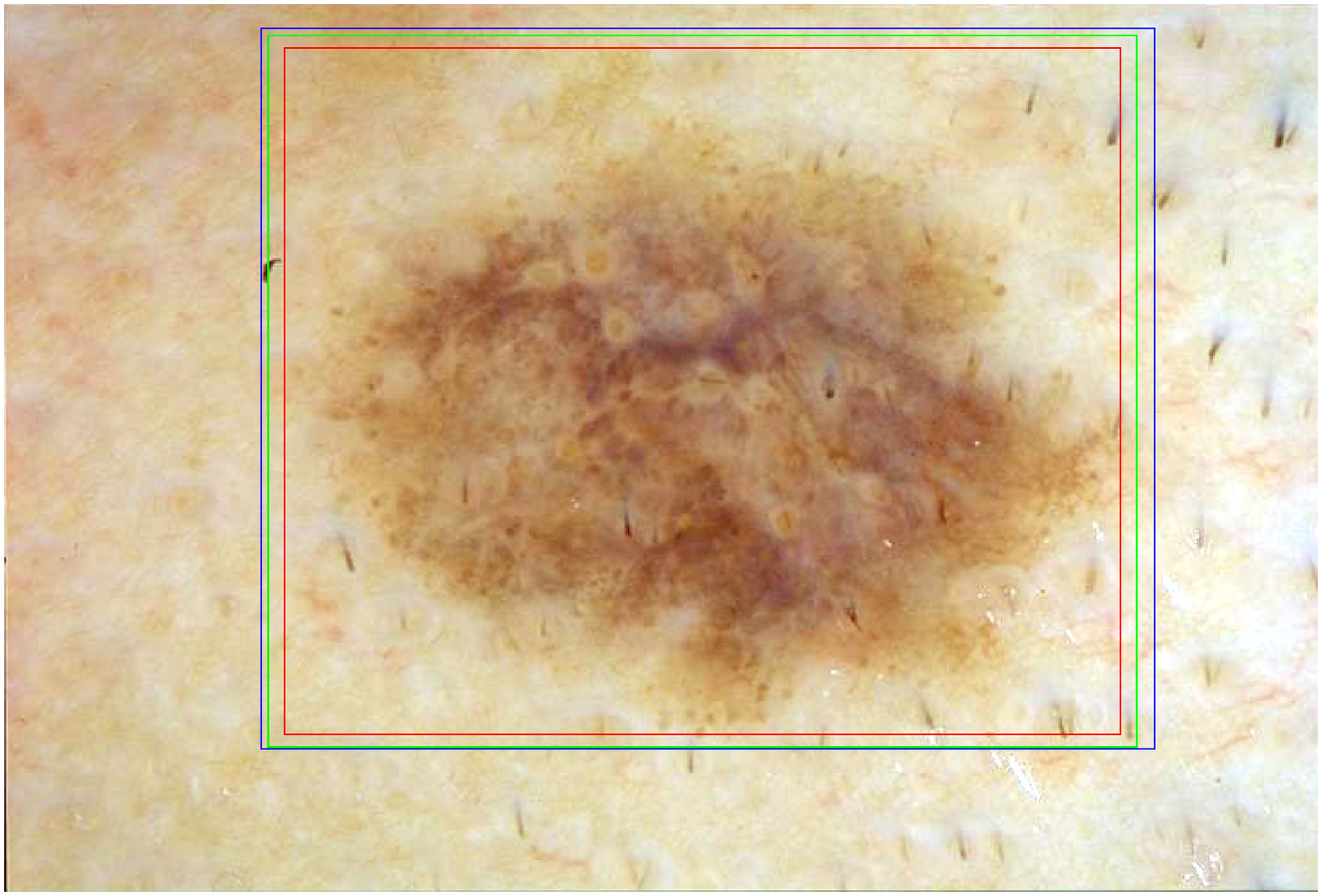}}
 \hspace{.2in}
 \subfigure[$\varepsilon_i = 3.18\% \mbox{\; , \;} \varepsilon_x = 4.82
\%$]{\label{results_d}\includegraphics[width=0.4\columnwidth]{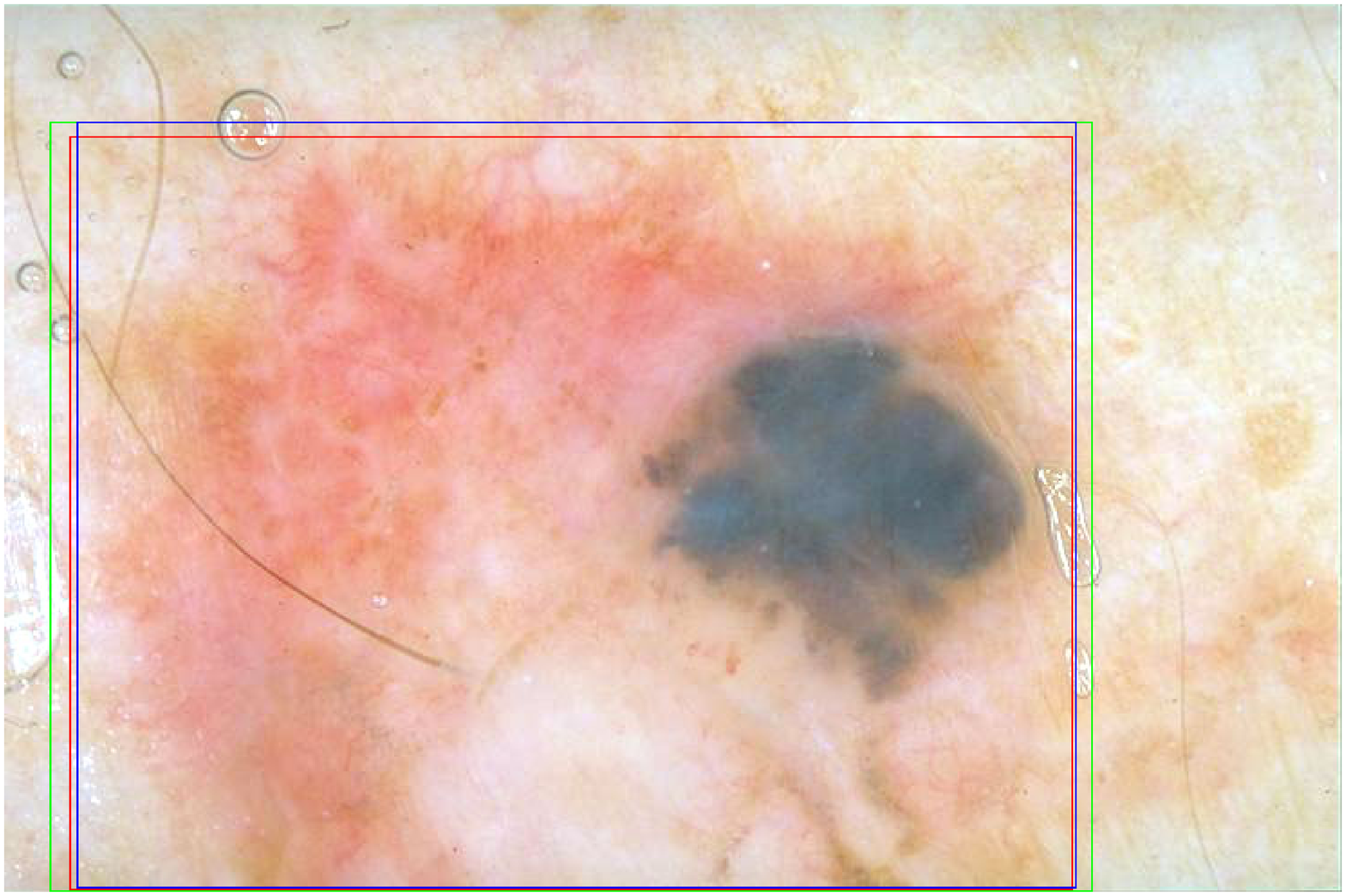}}
 \subfigure[$\varepsilon_i = 6.75\% \mbox{\; , \;} \varepsilon_x = 5.36
\%$]{\label{results_e}\includegraphics[width=0.4\columnwidth]{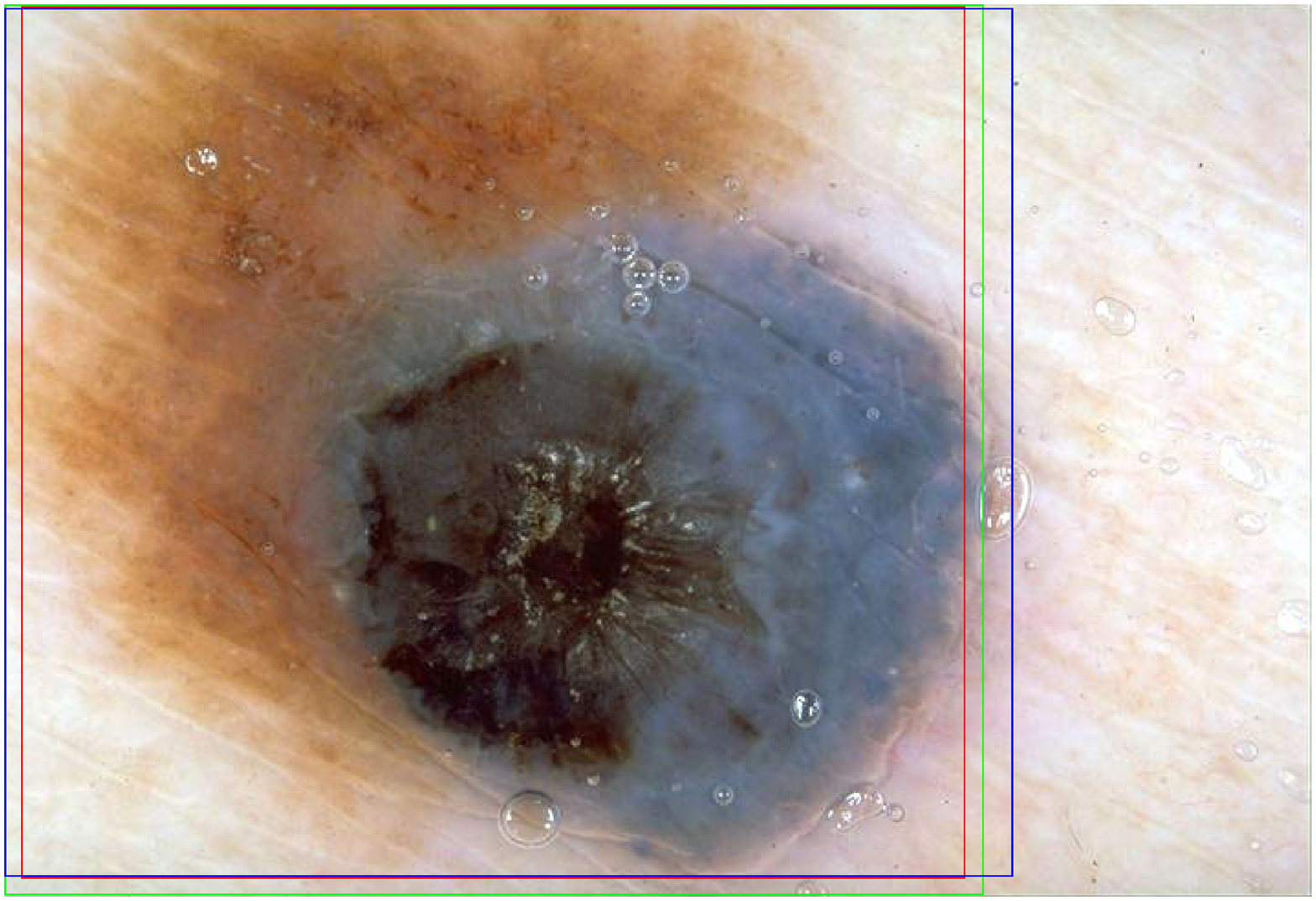}}
 \hspace{.2in}
 \subfigure[$\varepsilon_i = 14.24\% \mbox{\; , \;} \varepsilon_x = 9.21
\%$]{\label{results_f}\includegraphics[width=0.4\columnwidth]{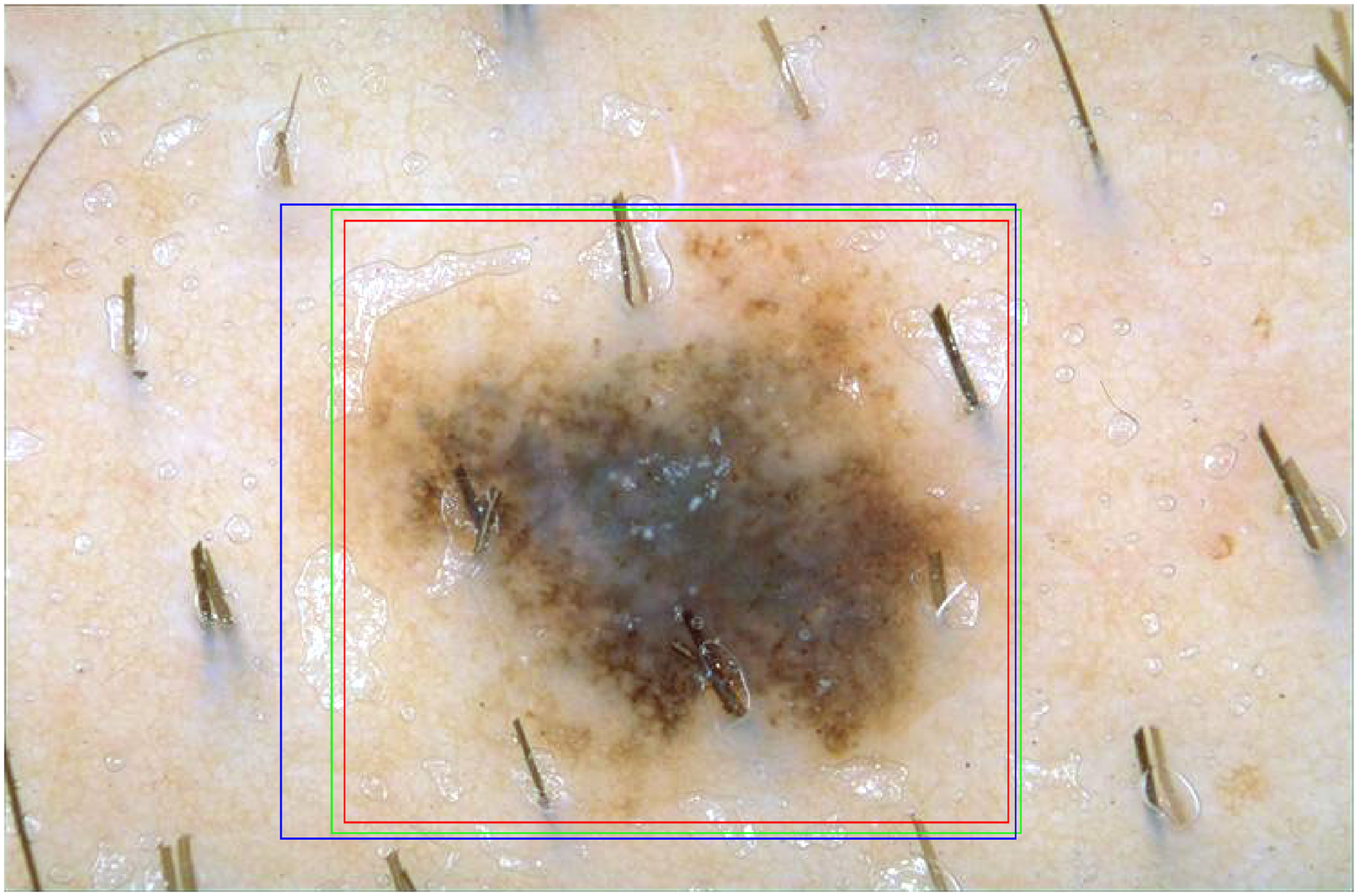}}
 \subfigure[$\varepsilon_i = 14.05\% \mbox{\; , \;} \varepsilon_x = 10.91
\%$]{\label{results_g}\includegraphics[width=0.4\columnwidth]{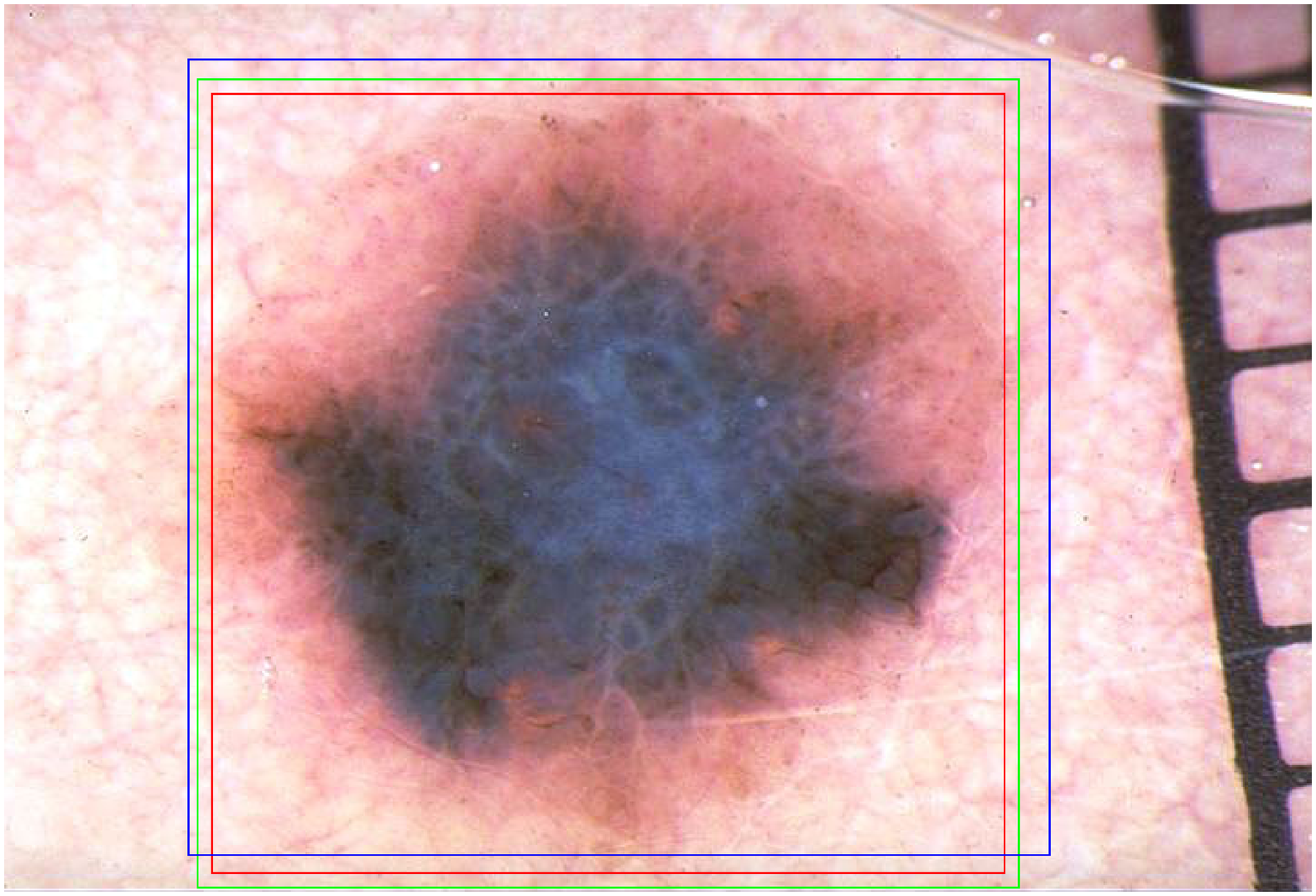}}
 \hspace{.2in}
 \subfigure[$\varepsilon_i = 18.36\% \mbox{\; , \;} \varepsilon_x = 13.62
\%$]{\label{results_h}\includegraphics[width=0.4\columnwidth]{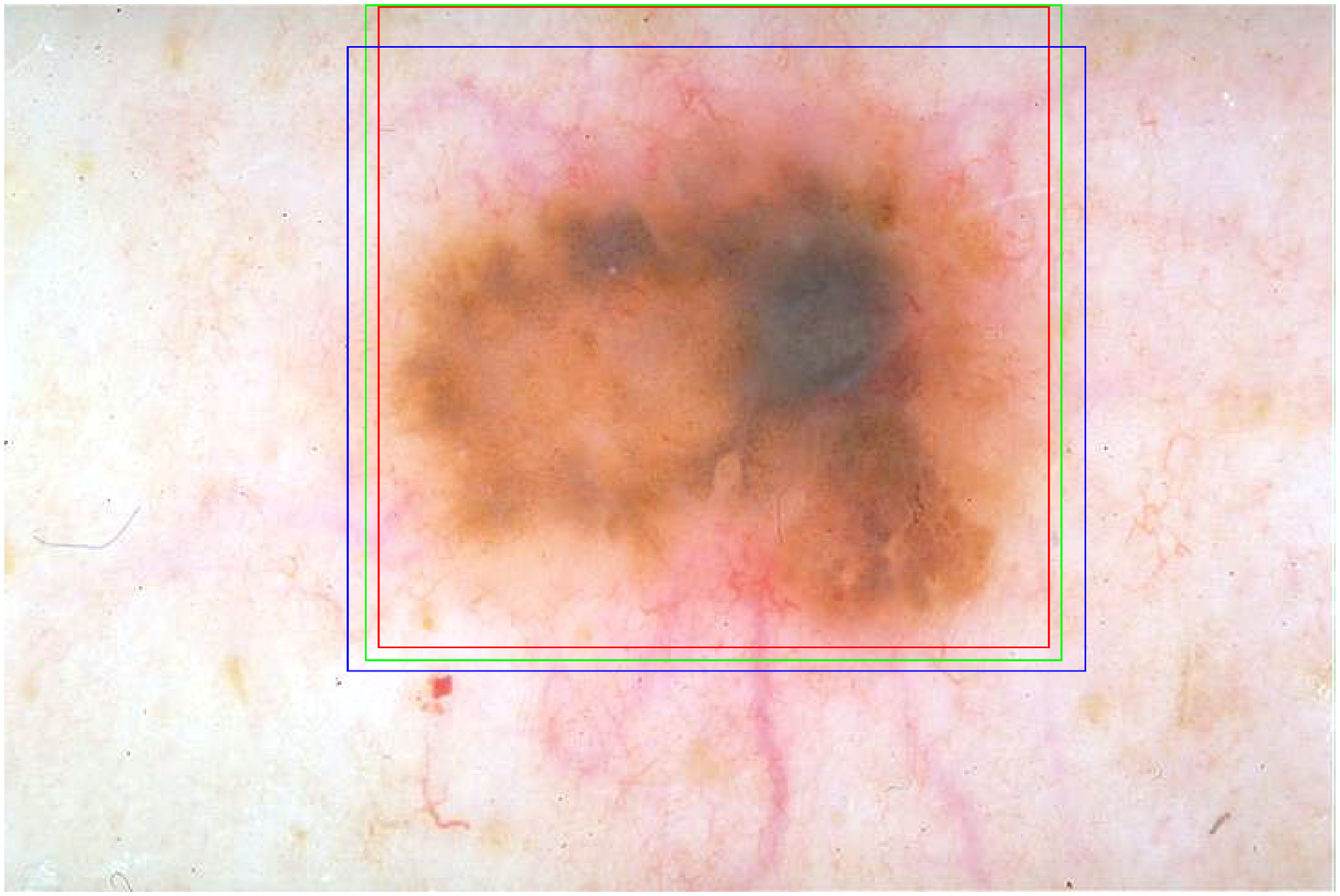}}
 \caption{Sample bounding box computation results ($\varepsilon_i$: initial box error, $\varepsilon_x$: expanded box error)}
 \label{sample_results}
\end{figure}

Fig.\ \ref{sample_results} shows sample bounding box computation results obtained using the ensemble Otsu-Kapur-Huang-Sahoo with $P = 2$. It can be seen that the presented method determines an accurate bounding box even for lesions with fuzzy borders. We note that while the expansion operation is useful in most cases, in some cases such as Fig.\ \ref{results_d}, it might deteriorate the results slightly.

\section{Conclusions}
In this paper, an automated method for approximate lesion localization in dermoscopy images is presented. The method is comprised of three main phases: black frame removal, initial bounding box computation using an ensemble of thresholding algorithms, and expansion of the initial bounding box. The execution time of the method is about 0.15 seconds for a typical image of size $768 \times 512$ pixels on an Intel Pentium D 2.66Ghz computer.
\par
The presented method may not perform well on images with significant amount of hair or bubbles since these elements alter the histogram, which in turn results in biased threshold computations. For images with hair, a preprocessor such as \emph{DullRazor\texttrademark} \cite{Lee97} might be helpful. Unfortunately, the development of a reliable bubble removal method remains an open problem.
\par
Future work will be directed towards testing the utility of the presented method in a border detection study. The implementation of the threshold fusion method will be made publicly available as part of the Fourier image processing and analysis library, which can be downloaded from \href{http://sourceforge.net/projects/fourier-ipal}{http://sourceforge.net/projects/fourier-ipal}

\section*{Acknowledgments}
This work was supported by grants from Louisiana Board of Regents (LEQSF2008-11-RD-A-12) and NIH (SBIR \#2R44 CA-101639-02A2).

\end{document}